%% file: main.tex
\documentclass[manuscript,screen, oneside]{acmart}
\acmJournal{THRI} 
\usepackage{caption}
\usepackage{subcaption}
\usepackage{chngcntr}
\AtBeginDocument{%
  \providecommand\BibTeX{{%
    \normalfont B\kern-0.5em{\scshape i\kern-0.25em b}\kern-0.8em\TeX}}}

\setcopyright{acmcopyright}
\copyrightyear{2024}
\acmYear{2024}
\acmDOI{XXXXXXX.XXXXXXX}

\acmConference[THRI]{Transactions on Human-Robot Interaction}
\acmBooktitle{Transactions on Human-Robot Interaction}
\acmISBN{978-1-4503-XXXX-X/18/06}

\usepackage{graphicx}
\usepackage{booktabs}
\usepackage{tabularx}

\begin{document}

\title[Consistency Matters: Defining Demonstration Data Quality Metrics]
{Consistency Matters: Defining Demonstration Data Quality Metrics in Robot Learning from Demonstration}
\author{Maram Sakr}
\affiliation{%
  \institution{University of British Columbia}
  \country{Canada}}

  \author{H.F. Machiel Van der Loos}
\affiliation{%
  \institution{University of British Columbia}
  \country{Canada}}

  \author{Dana Kuli{\'c}}
\affiliation{%
  \institution{Monash University}
  \country{Australia}}
  
\author{Elizabeth Croft}
\affiliation{%
  \institution{University of Victoria}
  \country{Canada}
}

\renewcommand{\shortauthors}{Sakr et al.}


\begin{abstract}

\input{Text/abstract}

\end{abstract}

\maketitle

\section{INTRODUCTION}
        \label{sec:introduction}
        \input{Text/introduction}

\section{RELATED WORK}
        \label{sec:related_works}
        \input{Text/related_works}

\section{PROPOSED APPROACH}
        \label{sec:approach}
        \input{Text/approach}

\section{EXPERIMENTAL DESIGN}
        \label{sec:exp}
        \input{Text/experiment}

\section{HYPOTHESES}
        \label{sec:hypo}
        \input{Text/hypotheses}


\section{RESULTS}
        \label{sec:results}

\input{Text/results}

\section{DISCUSSION}
        \label{sec:discussion}
        \input{Text/discussion}

\section{CONCLUSION AND FUTURE WORK}
        \label{sec:conclusion}
        \input{Text/conclusion}


\bibliographystyle{ACM-Reference-Format}
\input{main.bbl}

\appendix
    \section*{Appendix}
    \label{sec:appendix}

\input{Text/appendix}  
    \renewcommand{\thesection}{A.\arabic{section}} 
    \renewcommand{\thetable}{A\arabic{table}} 
    \setcounter{section}{0} 
    \setcounter{table}{0} 

\end{document}

%% file: Text/abstract.tex
Learning from Demonstration (LfD) empowers robots to acquire new skills through human demonstrations, making it feasible for everyday users to teach robots. However, the success of learning and generalization heavily depends on the quality of these demonstrations. Consistency is often used to indicate quality in LfD, yet the factors that define this consistency remain underexplored. In this paper, we evaluate a comprehensive set of motion data characteristics to determine which consistency measures best predict learning performance. By ensuring demonstration consistency prior to training, we enhance models' predictive accuracy and generalization to novel scenarios. We validate our approach with two user studies involving participants with diverse levels of robotics expertise. In the first study (\textit{N} = 24), users taught a PR2 robot to perform a button-pressing task in a constrained environment, while in the second study (\textit{N} = 30), participants trained a UR5 robot on a pick-and-place task. Results show that demonstration consistency significantly impacts success rates in both learning and generalization, with 70\% and 89\% of task success rates in the two studies predicted using our consistency metrics. Moreover, our metrics estimate generalized performance success rates with 76\% and 91\% accuracy. These findings suggest that our proposed measures provide an intuitive, practical way to assess demonstration data quality before training, without requiring expert data or algorithm-specific modifications. Our approach offers a systematic way to evaluate demonstration quality, addressing a critical gap in LfD by formalizing consistency metrics that enhance the reliability of robot learning from human demonstrations.

%% file: Text/introduction.tex
Learning from Demonstration (LfD), also known as imitation learning (IL), empowers robots to acquire new skills and behaviors by observing human demonstrations. This approach opens possibilities for everyday users, even those without robotics or programming expertise, to teach robots new tasks~\cite{argall2009survey}. However, while LfD holds significant promise, most algorithms implicitly assume that demonstrators are experts providing near-perfect demonstrations~\cite{ravichandar2020recent}. This idealized expectation often leads to a gap between real-world data quality and algorithmic assumptions, as practical demonstrations frequently include variability due to factors such as fatigue, lack of experience, or environmental constraints~\cite{amershi2014power, calinon2007teacher}.

Prior work has sought ways to address this gap by refining algorithms to cope with imperfect data~\cite{wu2019imitation, tangkaratt2020variational}. Some studies use heuristic quality indicators, such as undesired motions~\cite{osa2018algorithmic}, varying lengths and amplitudes~\cite{xu2024learning}, distribution shift~\cite{belkhale2024data}, or ambiguous demonstrations~\cite{sena2020quantifying}. However, these notions of quality are often vague and incomplete, lacking well-defined metrics to identify what makes a demonstration ``better`` or to systematically detect imperfections. This lack of systematic and coherent metrics creates challenges for researchers and practitioners alike, as inconsistent or suboptimal demonstrations can degrade learning performance and limit generalization~\cite{argall2009survey, osa2018algorithmic}.

Data quality can be assessed across various dimensions, each representing a measurable property of the data that influences its overall usefulness. This aligns with the broader definition of data quality in the ISO/IEC 25012 standard~\cite{ISO13586}, which is defined as \textit{``the degree to which a set of characteristics of the data meets the requirements``}. Examples of such characteristics in the machine learning literature include accuracy, completeness, consistency, and timeliness~\cite{cichy2019overview, laranjeiro2015survey}. Nevertheless, these general characteristics need to be adapted with domain-specific metrics and methods for measurement, particularly for LfD tasks. Questions arise, such as: \textit{How is accuracy defined in demonstrations? What constitutes consistency of the demonstrations?} 

\begin{figure*}[t]
\centering
\includegraphics[width=\textwidth]{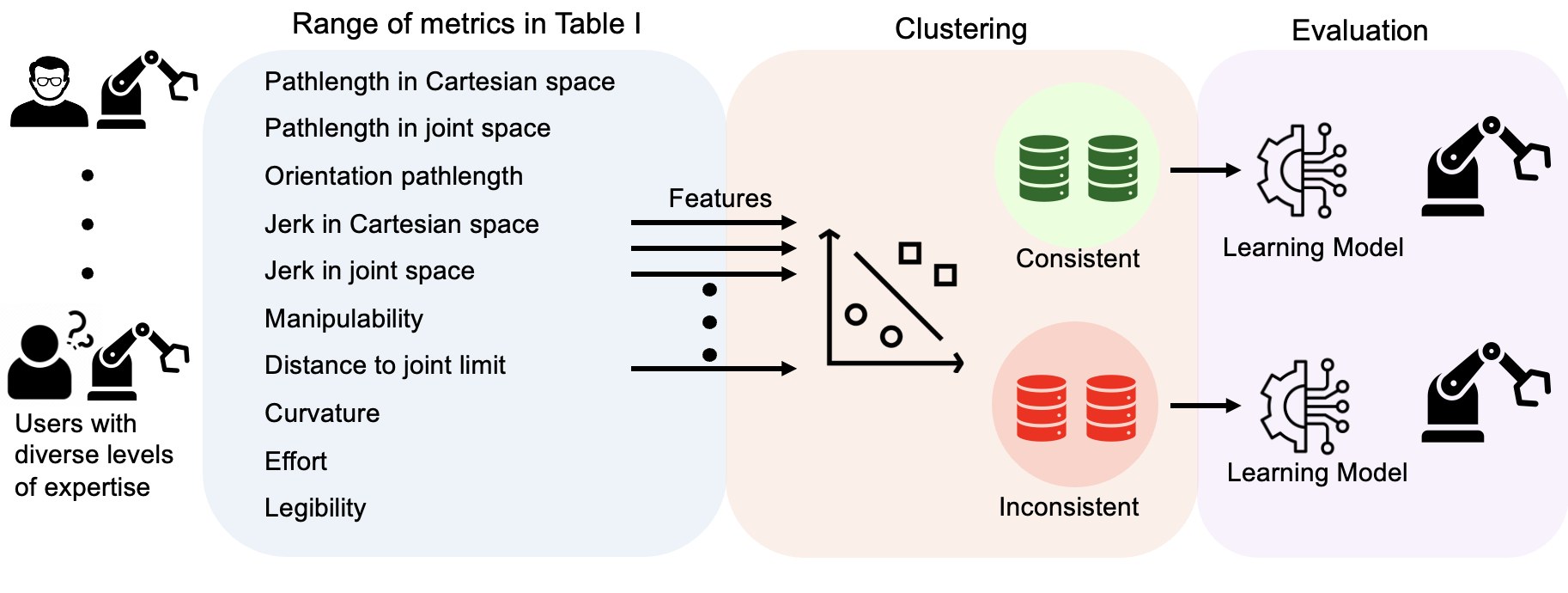}
\caption{Overview of the proposed approach. Data was collected from users with varying levels of expertise in robotics. The range of metrics listed in Table~\ref{tab:metrics} was calculated to serve as feature inputs for the clustering algorithm. The data was then clustered into two groups: consistent and inconsistent. A learning model was subsequently trained using consistent and inconsistent data. Finally, performance was evaluated by calculating the success rate of the trained models. }
\label{fig:overview}
\end{figure*}

Our approach, as shown in Fig.~\ref{fig:overview}, directly addresses this need by introducing a comprehensive set of metrics for defining and quantifying consistency in demonstration data, across both Cartesian and joint-space characteristics. By rigorously quantifying data quality in this way, we can ensure that only high-quality demonstrations feed into learning models, enhancing the predictive accuracy of both learning and generalization performance. This approach is designed to be applicable across diverse LfD scenarios, making it a practical tool for both researchers and practitioners. By validating demonstration consistency as a key data quality component, we propose a new approach for evaluating and curating data before model training, bridging a significant gap in LfD research and offering a scalable solution for improving robot learning outcomes.


%% file: Text/related_works.tex
Obtaining optimal demonstrations for all tasks that a robot must learn is a significant challenge~\cite{ravichandar2020recent}. Identifying experts proficient in both the task domain and in robotics to ensure high-quality demonstrations tailored to task goals and the robot's constraints is difficult~\cite{osa2018algorithmic}. Consequently, mixed-quality demonstrations are commonly used in learning. Knowing the quality of demonstrations beforehand allows lower-quality data to be filtered out, while multimodal learning approaches can be employed to focus on the highest quality demonstrations~\cite{han2012data, wang2017robust, tangkaratt2019vild}. 

\subsection{Learning from Mixed-Quality Demonstrations}
Examining the effect of diverse-quality demonstrations, Mandlekar et al.~\cite{mandlekar2021matters} found that IL models trained exclusively on high-quality demonstrations from a single human achieved higher success rates than models trained on larger datasets from multiple demonstrators with varied levels of proficiency. Similarly, in prior work~\cite{sakr2022quantifying}, we showed that models trained on mixed-quality data perform worse than those trained on a smaller dataset of consistently high-quality demonstrations, underscoring the importance of ensuring high-quality data from each demonstrator before aggregation. Our work builds on this foundation by independently evaluating the quality of demonstrations for each user, thereby enabling a more nuanced understanding of how demonstration quality impacts robot learning and generalization.

\subsection{Metrics for Evaluating Demonstration Quality}
\subsubsection{Task-Independent Quality Metrics}
The literature presents multiple approaches to defining and assessing demonstration quality, although few papers propose standardized or comprehensive metrics. Chernova and Thomaz~\cite{chernova2014robot} suggested evaluating human input quality based on factors such as task skill, time taken, and consistency; however, they did not provide specific metrics to measure these aspects. Kaiser et al.~\cite{kaiser1995obtaining} identified three primary sources of sub-optimality in human demonstrations: unnecessary actions that do not aid the goal, incorrect actions that reduce demonstration utility, and unmotivated actions influenced by human sensory information that is unavailable to the robot. While they proposed methods to measure unnecessary actions via motion ratios and discontinuity checks, identifying incorrect actions remains challenging, and distinguishing motivated from unmotivated actions often depends on the demonstration method, such as teleoperation or kinesthetic teaching~\cite{billard2008survey}.

Other studies focused on teaching efficacy and error detection within demonstrations. Fischer et al.~\cite{fischer2016comparison} identified common user errors, such as excessive pressure on the gripper, proximity to singularities, and self-collisions, although they did not evaluate these errors' impact on robot learning and generalization. Sena and Howard~\cite{sena2020quantifying} addressed data sparsity by defining two quality metrics: teaching efficacy, which measures task-space generalization, and teaching efficiency, which normalizes efficacy by the number of demonstrations. These metrics focus on optimizing demonstration quantity rather than assessing individual demonstration quality, which remains critical for ensuring generalizable and reliable learning.

\subsubsection{Task-Dependent Quality Metrics}
Researchers have developed task-specific metrics for assessing demonstration consistency. Ureche and Billard~\cite{pais2015metrics} proposed metrics for bimanual tasks, such as evaluating tool maneuverability, consistency across trials, and coordination between arms. They measured consistency by tracking constraint changes across each dimension, although performance tends to vary over time, leading to fluctuations in quality. This limitation points to the need for trial-independent evaluations to capture high-quality demonstrations more effectively. Additionally, our work proposes an extensive set of motion features to determine which measures best predict learning and generalization performance.

Despite the recognized need for quality assessments, there is a notable lack of metrics and methods that comprehensively evaluate multiple dimensions of demonstration quality. This paper addresses that gap by introducing task-agnostic metrics to assess consistency in demonstration data, focusing on motion characteristics in both Cartesian and joint-space dimensions. To validate these metrics, we conducted two user studies, demonstrating that our consistency metrics reliably predict both task success and generalization performance. The results highlight the robustness of the metrics in different tasks and robotic platforms. By applying these measures to evaluate demonstrations before learning, we can ensure the inclusion of only high-quality data, ultimately enhancing the efficiency and effectiveness of the Learning from Demonstration (LfD) process.

%% file: Text/approach.tex

We begin by proposing a broad list of metrics to evaluate the consistency of demonstration data. These metrics evaluate various aspects of robot motion and investigate their influence on learning and generalization performance. These metrics consider both Cartesian and joint spaces for the robot. Table~\ref{tab:metrics} shows the proposed metrics for evaluating the consistency of the demonstration data. Path length metrics are selected to detect unnecessary motions~\cite{osa2018algorithmic} that do not contribute to the task's goal. For instance, if a user struggles to manoeuvre a robot toward the goal, the provided trajectories will be longer than those of a proficient user who manoeuvres the robot directly toward the goal. Additionally, trajectory smoothness improves motion stability and smoothness in the learned and generalized trajectories~\cite{chen2003programing}. Fluent trajectories that avoid abrupt changes are also important for efficient navigation, reducing wear and tear on robotic components, and minimizing energy consumption~\cite{ryuh1989robot}. Hence,  we propose jerk, curvature, and effort as potential metrics to quantify the quality and consistency of the demonstrations. 

The robot's posture directly impacts reaching movements and manipulation tasks (e.g., pushing, pulling, reaching)~\cite{jaquier2021geometry}. In this context, manipulability~\cite{yoshikawa1985manipulability} serves as a kinematic descriptor indicating the ability to arbitrarily move in different directions of the task in a given joint configuration. Since the ability to manoeuvre the robot in the workspace degrades at singular configurations, this quality metric is important for allowing the robot to generalize the learned task across the workspace~\cite{vahrenkamp2015representing, Jaquier2020geometry}. In addition to singularities, joint limits significantly affect the manoeuvrability of the robot in the workspace~\cite{flacco2012prioritized}.

If the robot operates in the same space as humans, it is important to consider legibility as a quality metric~\cite{dragan2013legibility}. Legibility measures how easily an observer can infer a robot's goal from its motion. It is defined based on the probability assigned to the actual goal across the trajectory. This probability is computed using the cost function for each potential goal, which combines the early differentiation of the trajectory and the progress towards the goal~\cite{wallkotter2022new}. Trajectories with high legibility will allow an observer to infer the goal early, given a small segment of the trajectory, and all points in the trajectory will be directed toward that goal. Details on the legibility calculation are in Appendix~\ref{appendix: leg}.

\begin{table*}[t]
    \centering
    \renewcommand{\arraystretch}{1.8}
    \caption{Proposed quality metrics. The variables denoted here are: $x$ for end-effector position, $q$ for joint angle, $R$ for rotation matrix, $\ddot{x}$ for end-effector acceleration, $\dddot{x}$ for end-effector jerk, $\dddot{q}$ for joint jerk, $\tau$ for torque, $J$ for the Jacobian, $n_d$ for the total number of degrees of freedom (DOFs), $q_d^-$ for the lower limit of the $d^{th}$ joint, $q_d^+$ for the upper limit of the $d^{th}$ joint. \( P(G_i | \xi_{1:t}) \) is the posterior probability of goal \( G_i \) given the trajectory segment \( \xi_{1:t} \).}
    \begin{tabularx}{\textwidth}{XlX}
    \toprule
    \textbf{Criterion} & \textbf{Reference name} & \textbf{Formula} \\
    \midrule
    
        Path Length in Cartesian Space & $Q_{x}$ & $\sum_{t=1}^{T}\|x_{t} - x_{t-1}\|^2$ \\
        Path Orientation Length &  $Q_{rot}$ & $\sum_{t=1}^{T} \arccos\left(\frac{\text{trace}(R_{t} \cdot R_{t-1}^T)}{2}\right)^2$ \\
        Path Length in Joint Space &  $Q_{q}$ & $\sum_{t=1}^{T}\|q_{t} - q_{t-1}\|^2$ \\
        Jerk in Cartesian Space &  $Q_{\dddot{x}}$ & $\sum_{t=1}^{T}{\dddot{x}_{t}}^2$ \\
        Jerk in Joint Space &  $Q_{\dddot{q}}$ & $\sum_{t=1}^{T}{\dddot{q}_{t}}^2$ \\
        Manipulability &  $Q_{M}$ & $ \sum_{t=1}^{T} \sqrt{\det(J_t J^T_t)}$ \\
        Distance to Joint Limit &  $Q_{q_{lim}}$ & $ \sum_{t=1}^{T} \prod_{d}^{n_d} \frac{4(q_d - q_d^-) (q_d^+ - q_d)}{(q_d^+ - q_d^-)^2}$ \\
        Joint Effort &  $Q_{\tau}$ & $ \sum_{t=1}^{T} \sum_{d}^{n_d} \tau_{d,t}^2$ \\
        Cartesian Curvature & $Q_{\kappa}$ & $\sum_{t=1}^{T} \frac{\|\dot{x}(t) \times \ddot{x}(t)\|}{\|\dot{x}(t)\|^3}$ \\
        Legibility & $Q_{L}$ & $-\sum_{i=1}^{n} P(G_i | \xi_{1:t}) \log P(G_i | \xi_{1:t})$ \\
        \bottomrule
    \end{tabularx}
    \label{tab:metrics}
\end{table*}

It is worth noting that task goals are the primary objectives of the demonstrator, with quality being a secondary concern. For instance, when manoeuvring a robot in a constrained space, avoiding collisions takes precedence over the robot's manipulability or proximity to joint limits. Thus, it is essential to consider task and workspace constraints when applying these metrics. To reduce the dependency on task-specific constraints, we focus on the \textit{consistency} of demonstrations in terms of these metrics rather than their absolute values.  

Consistency allows us to account for variations in task difficulty and environmental constraints. Furthermore, consistent demonstrations indicate a proficient demonstrator who can repeatedly perform tasks efficiently and effectively~\cite{pais2015metrics}. This proficiency is more informative for learning than absolute performance metrics~\cite{beliaev2022imitation}, which can be skewed by a single good or bad demonstration. By quantifying the consistency of demonstrations, researchers can prioritize the most consistent ones for learning, ultimately ensuring improved learning and generalization performance.

%% file: Text/experiment.tex
To validate that the consistency of the proposed metrics is predictive of demonstration quality and improves learning performance (as in Fig.~\ref{fig:overview}), we collected demonstrations from users with diverse robotics experiences using kinesthetic teaching. The demonstrations were then clustered into consistent and inconsistent groups based on the proposed metrics. Finally, we evaluated the learning and generalization performance of both groups to examine the correlation between the metrics and the robot's performance. Kinesthetic teaching allows users to physically guide the robot to achieve a task, with the robot's state recorded via its onboard sensors (e.g., joint angles, torques)~\cite{argall2009survey}. Recording demonstrations directly on the robot using its integrated sensors eliminates the correspondence problem, which arises from the mismatch between the human teacher and the robot learner due to differences in sensing ability, body structure, and mechanics~\cite{chernova2014robot}. We experimentally validated the proposed approach across two different tasks and robots to demonstrate the generalizability of the proposed metrics. 

\subsection{Experiment 1: Button Pressing}
\subsubsection{Task Definition}

\begin{figure*}[t]
    \centering
    \begin{subfigure}[t]{0.24\textwidth}
        \centering
        \includegraphics[width=\textwidth]{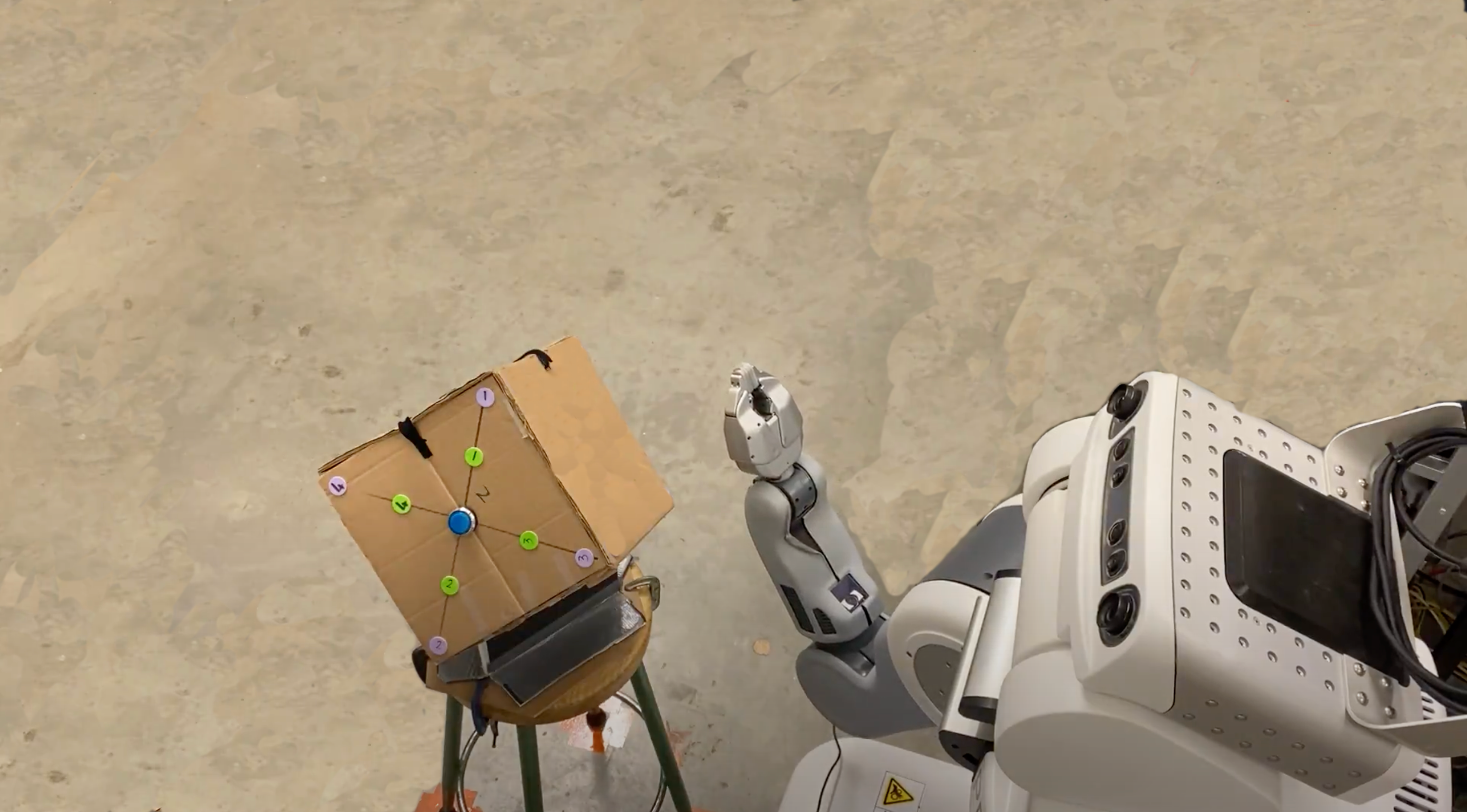}
        \caption{(a) Robot's initial position}
        \label{fig:exp_initial}
    \end{subfigure}
    \hfill
    \begin{subfigure}[t]{0.24\textwidth}
        \centering
        \includegraphics[width=\textwidth]{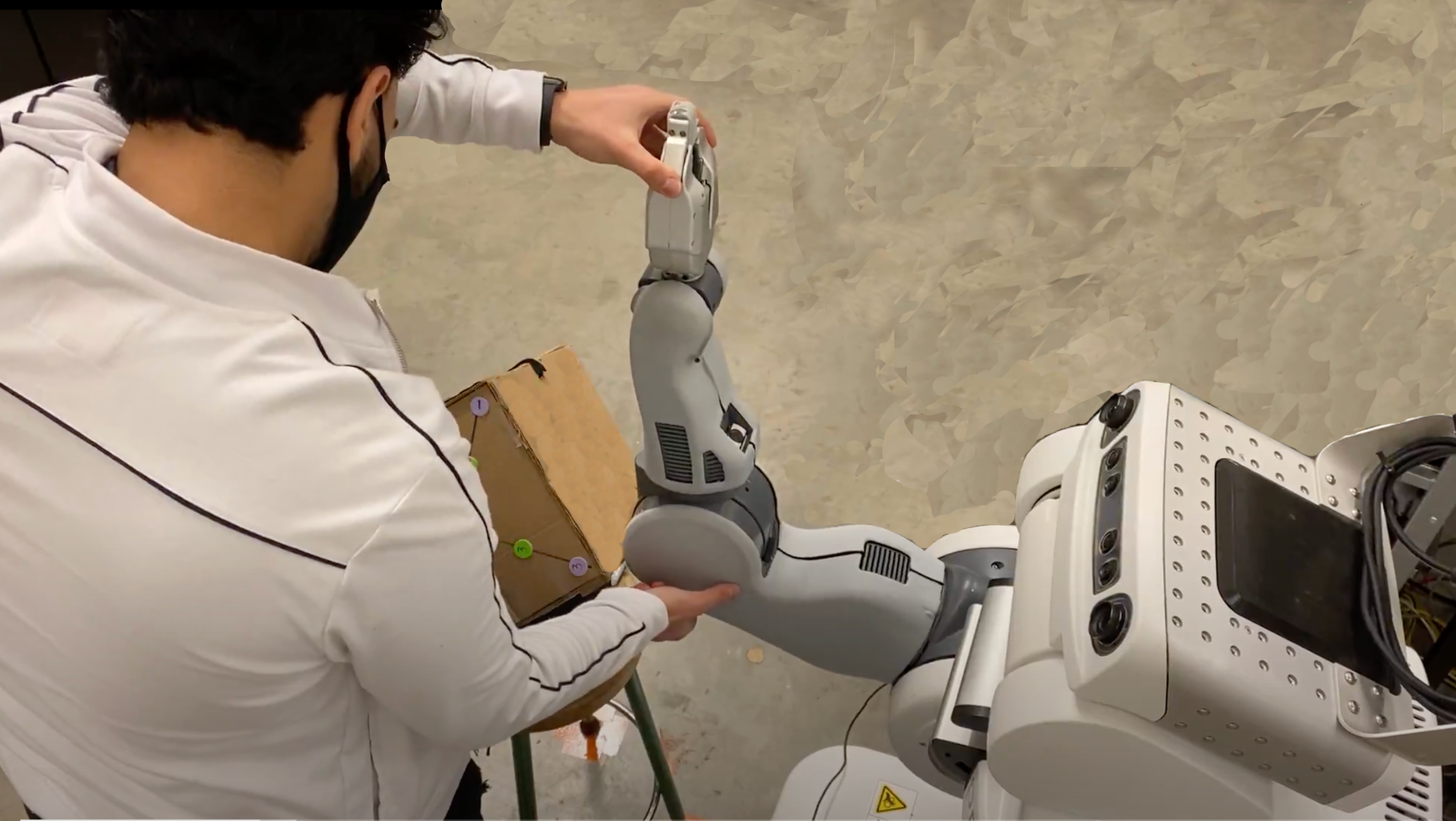}
        \caption{(b) Struggling to manoeuvre around the box}
        \label{fig:exp_struggle}
    \end{subfigure}
    \hfill
    \begin{subfigure}[t]{0.24\textwidth}
        \centering
        \includegraphics[width=\textwidth]{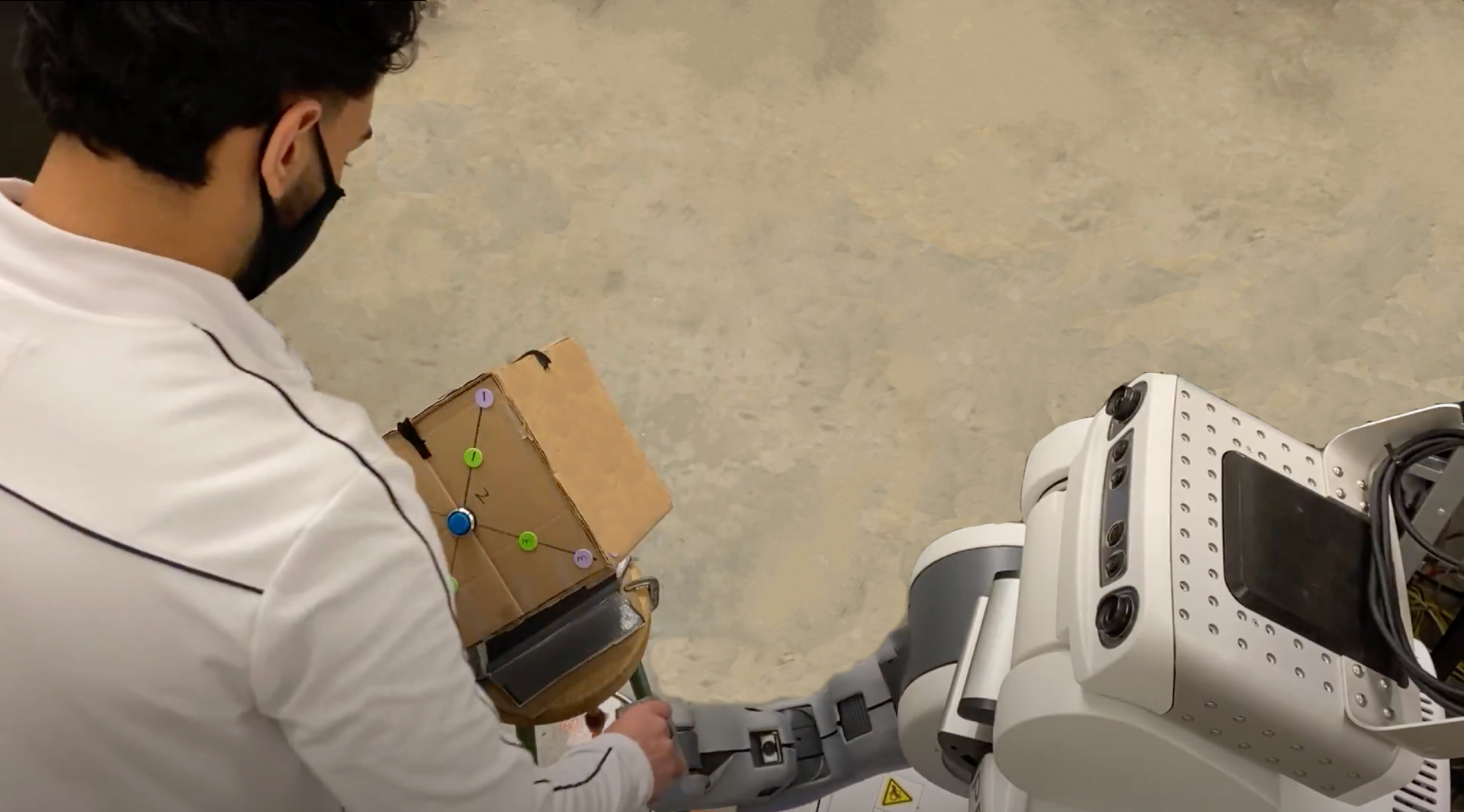}
        \caption{(c) Finding a better way to manoeuvre}
        \label{fig:exp_good}
    \end{subfigure}
    \hfill
    \begin{subfigure}[t]{0.24\textwidth}
        \centering
        \includegraphics[width=\textwidth]{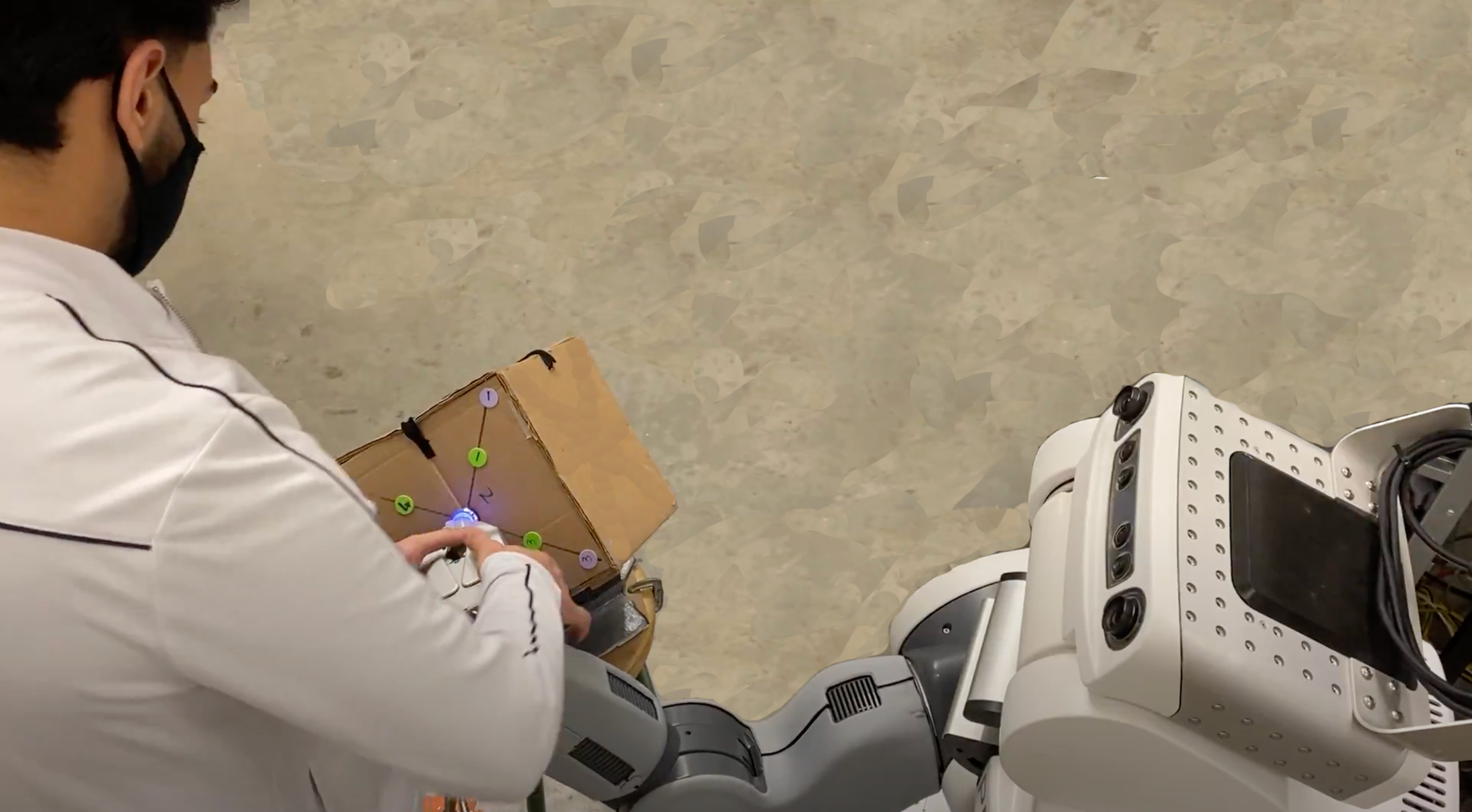}
        \caption{(d) Pressing the button}
        \label{fig:exp_final}
    \end{subfigure}
    \caption{Button-pressing task demonstration overview from (a) the initial position of the robot, (b) an example of a user's struggle to get the robot around the box, (c) an example of a better manoeuvre of the robot around the box, and (d) the robot pressing the button.} 
    \label{fig:experiment}
\end{figure*}

Our first exemplar task is pressing a button, a generic task that models real-world tasks such as pressing a doorbell, an elevator call button, or a pedestrian crossing button. This task was selected because it does not require domain expertise but does require practice with the robot to provide high-quality demonstrations. The task involves both a constrained reaching task and a fine control motion for pressing the button.

The robot platform used in this work is the PR2 (Willow Garage, Personal Robot 2), a mobile manipulator with two 7-DoF arms operating under a ROS Melodic. The passive spring counterbalance system in PR2’s arms provides gravity compensation, giving users the ability to kinesthetically move the robot's arms within their kinematic range. Each arm has a 1-DoF under-actuated gripper. In this experiment, we only used the right arm in gravity compensation mode with the gripper closed. 

Fig.~\ref{fig:experiment} shows the experimental setup used for data collection. A cardboard box was fixed on one of its vertices so that all buttons were reachable by the robot gripper. Only one face of the box was used in the data collection. Buttons were placed in the center (large blue button), corners (purple foam markers), and midway between the corners and the center of each face (green foam markers). A total of nine-goal positions were used. This setup represents a reaching task in a constrained space, requiring the participant to manoeuvre the robot arm around the box to reach the goal positions while avoiding self-collisions and collisions with the box.

\subsubsection{User Study}

We recruited participants through advertisements on the university campus and social media. A total of 24 participants (18 male, 6 female) with an average age of 22 years and varying levels of robotics expertise (ranging from no experience to six years or more) participated in the study. The University Behavioral Research Ethics Board approved the research (application ID H20-03740), and informed consent was obtained from each participant before the experiment began.

The experiment was conducted in two sessions on different days, acknowledging the importance of practice in motor skill learning~\cite{schmidt2008motor}. We aimed to explore which consistency metrics improved from the first to the second session and how this would be reflected in robot learning and generalization.  The experimenter briefly explained the long-term objective of the research and instructed the participants that their task was to program new skills into the PR2 robot. Participants were told to imagine a scenario in which they brought a robot to their home and wanted to teach it a task. The robot would imitate and learn from their demonstrations, so they should provide as natural demonstrations as possible. 

The robot was set in gravity compensation mode, and participants were asked to hold the robot's right arm and physically guide it to press a target button (kinesthetic teaching). The right arm always started in the same position, with the elbow at 90° and the gripper pointed up (untucked position), as in Fig.~\ref{fig:exp_initial}. After each demonstration, the experimenter teleoperated the right arm via a joystick to return it to the initial position. The participants demonstrated the task for three trials in each session. The robot's joint angles were recorded during each demonstration and saved as ROSbag files for offline analysis.

Each participant provided 27 demonstrations per session. After the first session, the participants reflected on their learning and practice and scheduled the second session on a different day. This procedure was motivated by Walker et al.~\cite{walker2003sleep}, who found that a night of sleep after motor skill training significantly improves skill levels in subsequent repetitions. In the second session, participants repeated the same procedure as in the first session. 

\subsection{Experiment 2: Pick and Place}

\begin{figure*}[t]
    \centering
    \begin{subfigure}[t]{0.22\textwidth}
        \centering
        \includegraphics[width=\textwidth]{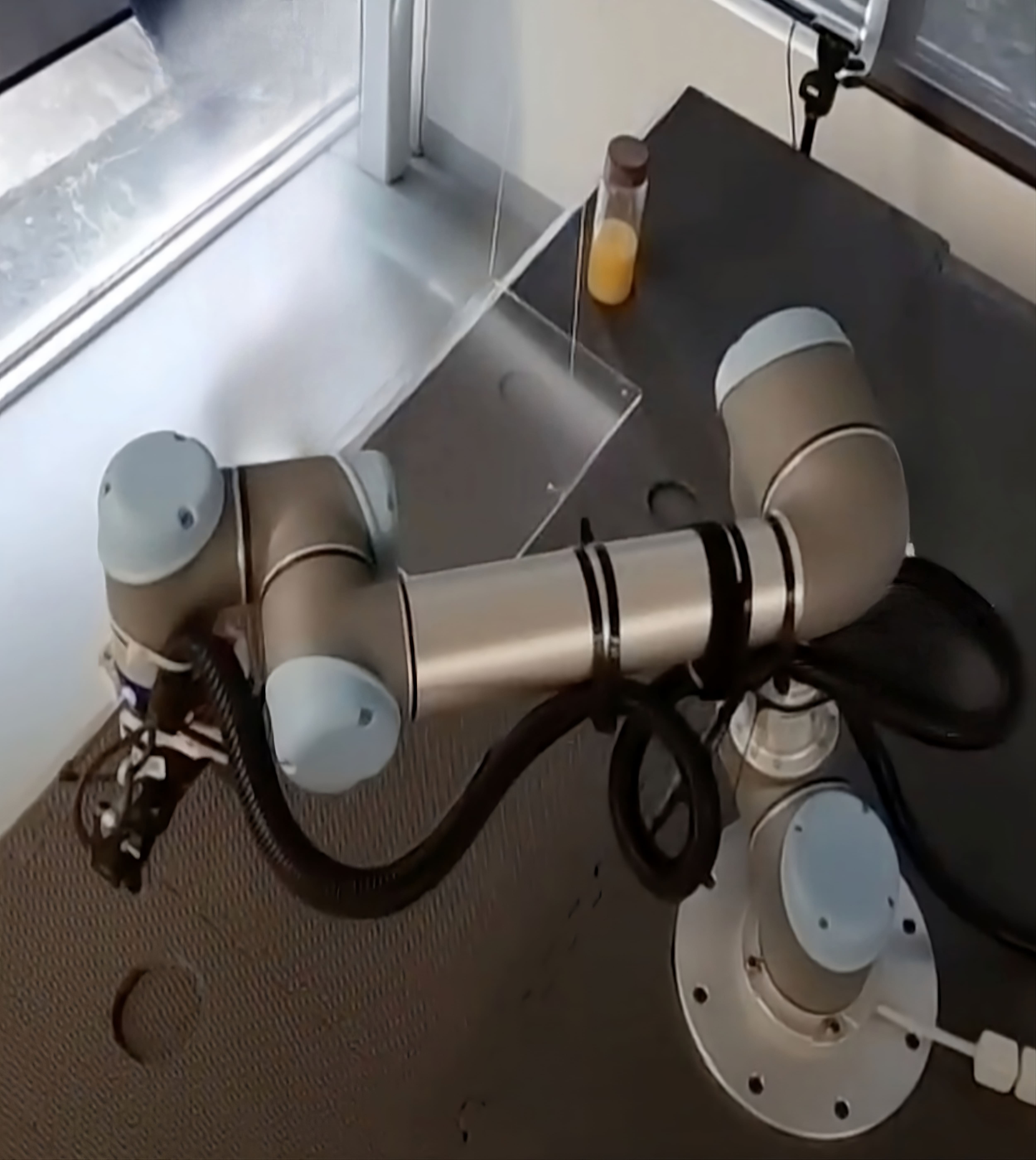}
        \caption{(a) Robot's initial position}
        \label{fig:exp2_initial}
    \end{subfigure}
    \begin{subfigure}[t]{0.22\textwidth}
        \centering
        \includegraphics[width=\textwidth]{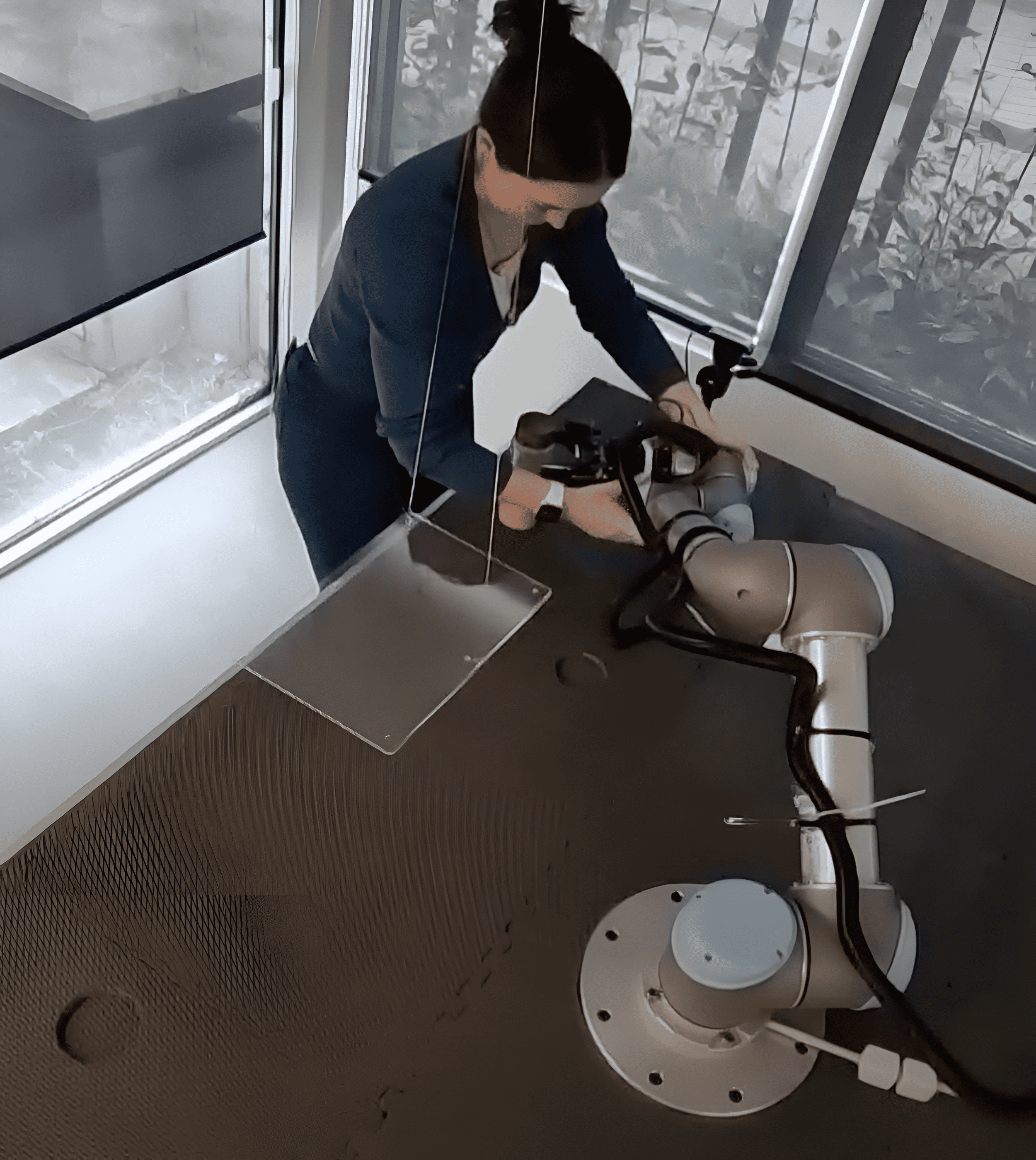}
        \caption{(b) Picking up the bottle}
        \label{fig:exp2_pickup}
    \end{subfigure}
    \begin{subfigure}[t]{0.22\textwidth}
        \centering
        \includegraphics[width=\textwidth]{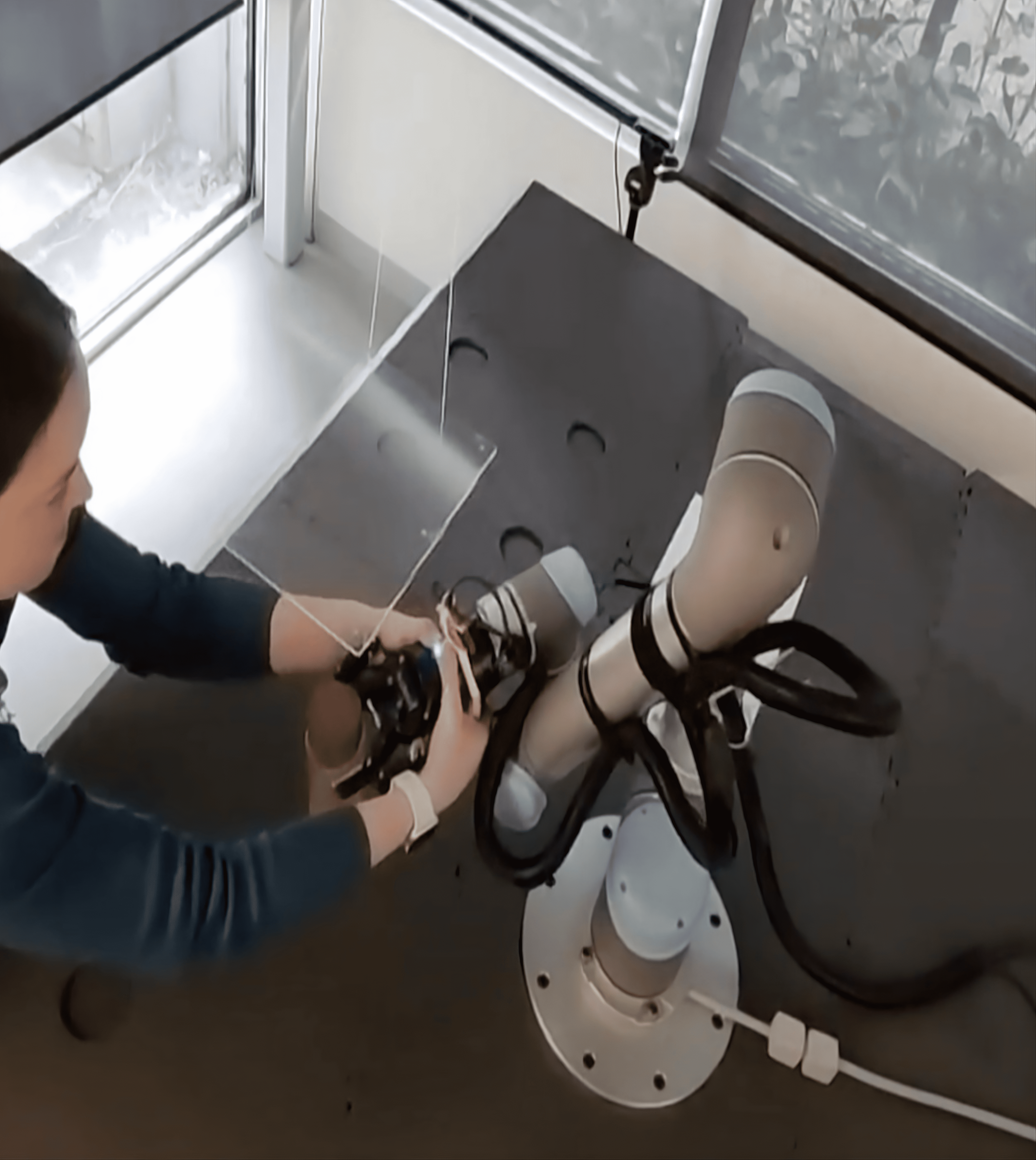}
        \caption{(c) manoeuvre the robot under the obstacle}
        \label{fig:exp2_barrier}
    \end{subfigure}
    \begin{subfigure}[t]{0.22\textwidth}
        \centering
        \includegraphics[width=\textwidth]{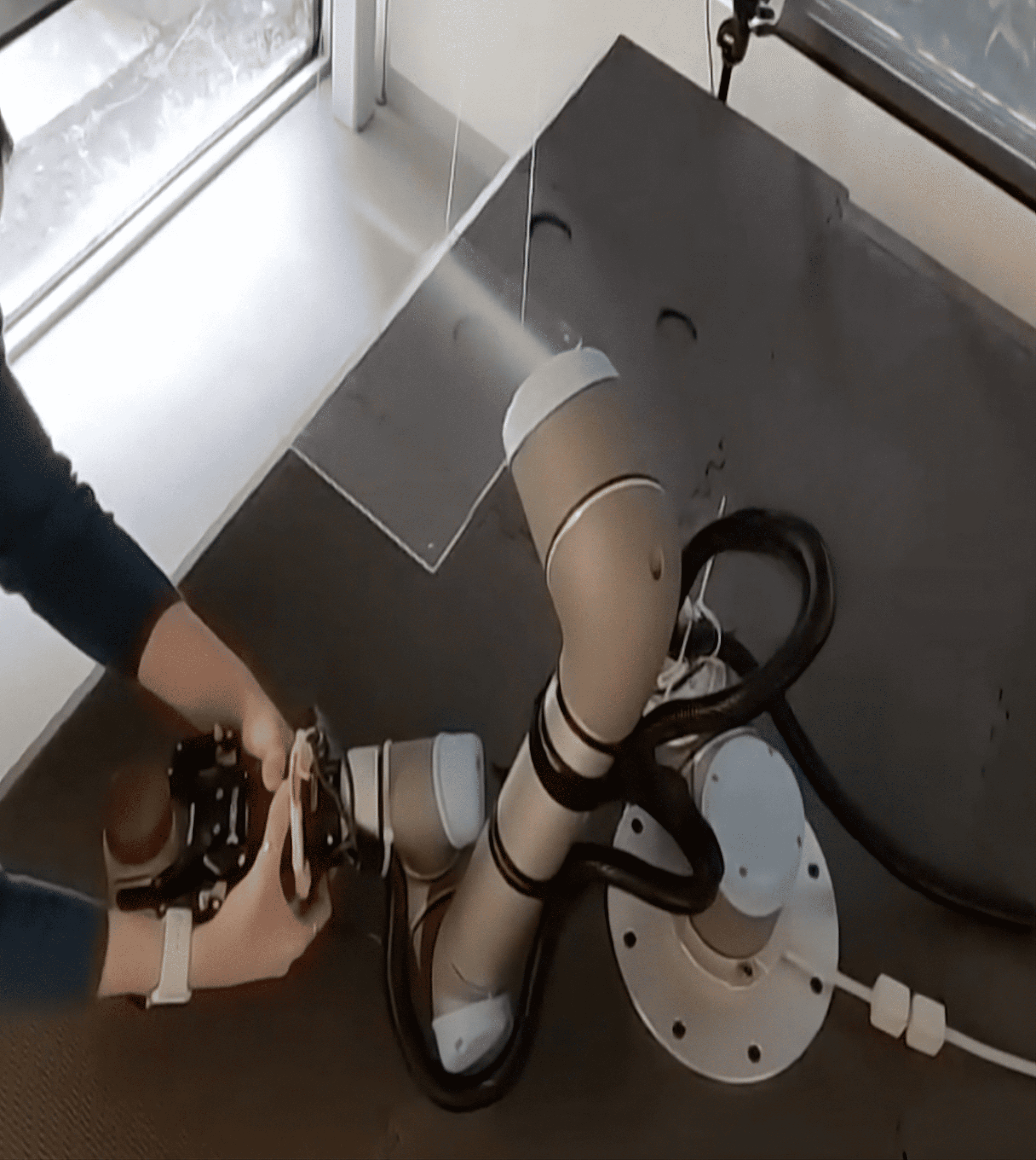}
        \caption{(d) Placing the bottle}
        \label{fig:exp2_place}
    \end{subfigure}
    
    \vspace{0.5em} 
    \begin{subfigure}[t]{0.4\textwidth}
        \centering
        \includegraphics[width=\textwidth]{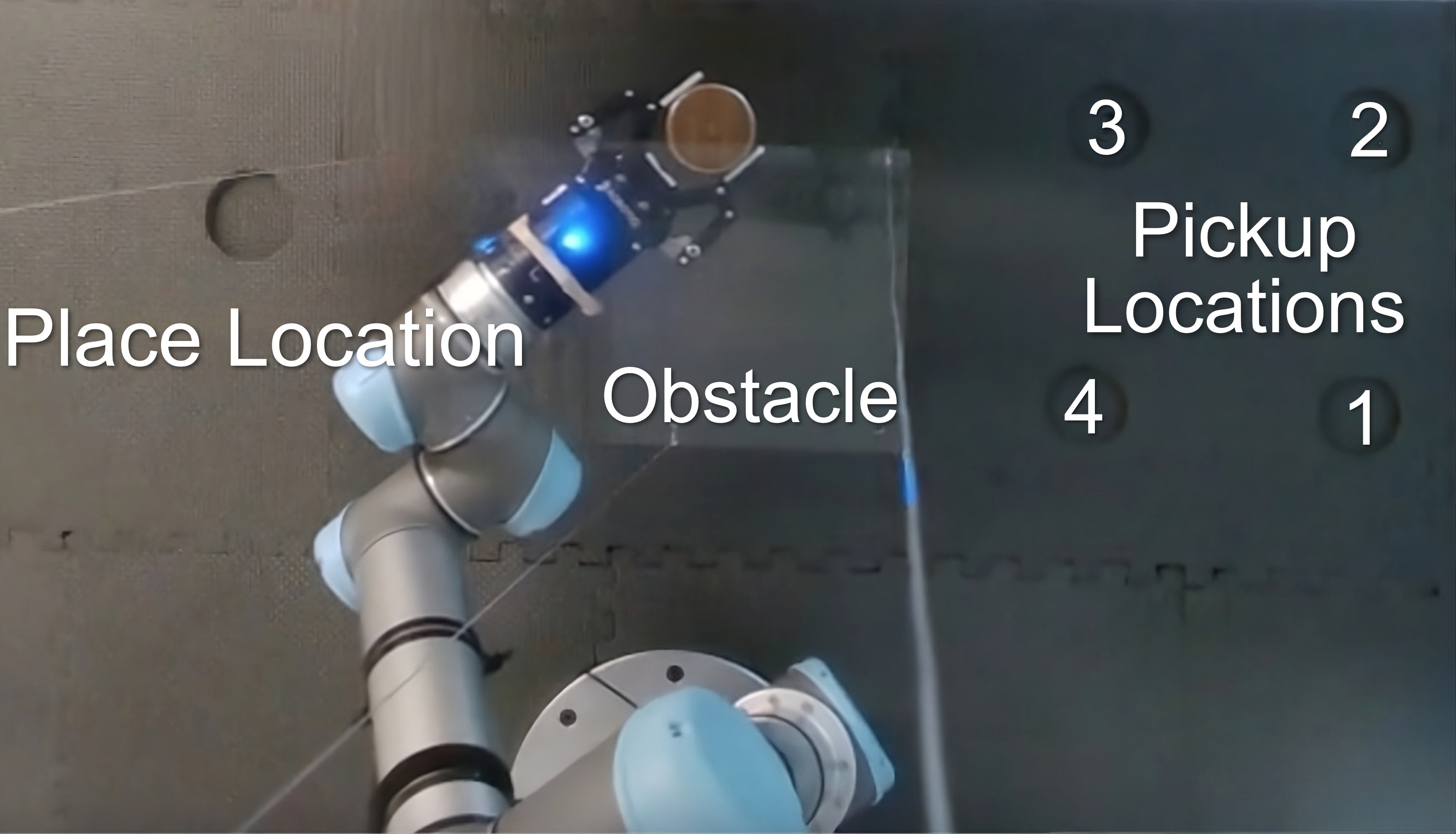}
        \caption{(e) Top view showing the pickup and place locations}
        \label{fig:exp2_top}
    \end{subfigure}
    \hspace{1em} 
    \begin{subfigure}[t]{0.17\textwidth}
        \centering
        \includegraphics[width=\textwidth]{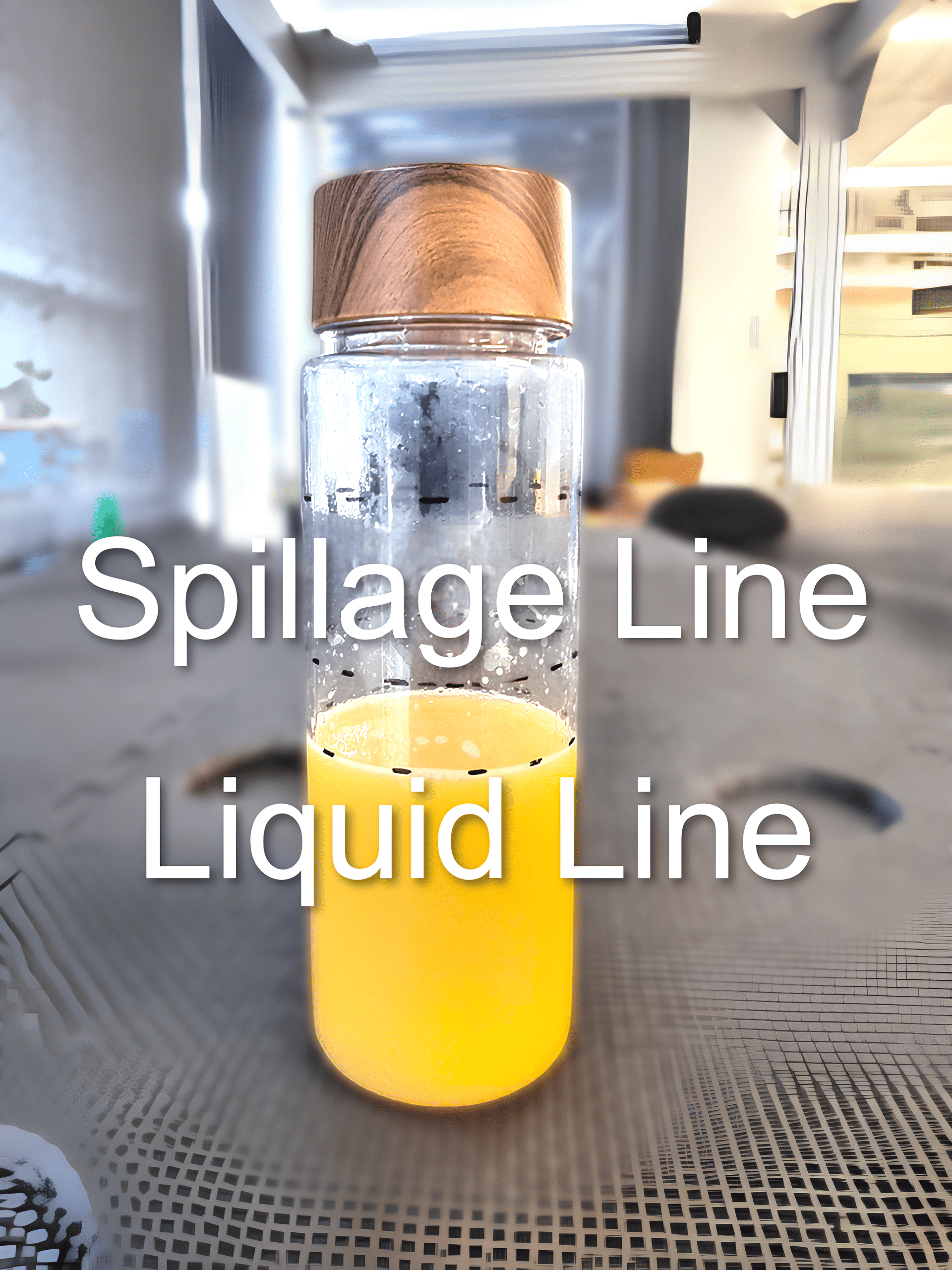}
        \caption{(f) The experimental bottle with spillage line}
        \label{fig:exp2_bottle}
    \end{subfigure}
    
    \caption{Task demonstration overview: (a) Robot's initial position, (b) Picking up the bottle, (c) manoeuvring under the obstacle, (d) Placing the bottle, (e) Top view of pickup, obstacle and place locations, and (f) Experimental bottle with spillage line.} 
    \label{fig:experiment2}
\end{figure*}

\subsubsection{Task Definition}
Our second exemplar experimental task represents picking and placing a liquid-filled container while avoiding collisions and spillage. This task does not require any domain expertise, only some practice to provide smooth demonstrations from which the robot can learn. Fig.~\ref{fig:experiment2} shows the experimental setup with a Universal Robots (UR5) equipped with a Robotiq 2F-85 two-finger gripper as its end effector and operating under ROS Melodic. The task is to pick up the bottle from any of the four positions in the pickup area and place it in the designated ``place`` location, as shown in Fig.~\ref{fig:exp2_top}, without ``spilling``. An obstacle is positioned above the table, midway between the pick and place locations. 

To simplify spillage detection, the robot picks up a sealed bottle with landmarks (dotted lines above the liquid level) to define the ``spillage`` threshold, as shown in Fig.~\ref{fig:exp2_bottle}. A 2 cm margin was set between the top surface of the liquid at rest and the spillage landmark. If the user significantly changes the end effector orientation, provides jerky trajectories, or moves the robot too quickly, the liquid may exceed the spillage mark.

\subsubsection{User Study}
Participants were recruited through social media and word of mouth. A total of 30 participants (24 male, 5 female, 1 non-binary) with an average age of 27 years took part in the study. The participants had varying expertise in robotics, with 33.3\% never having prior physical interaction with robots. The study was approved by the Monash University Human Ethics Committee (Project ID: 37303), and informed consent was obtained from each participant before the experiment began.

Participants were introduced to the experimental setup and the task. The experimenter demonstrated physically guiding the robot to perform the task and emphasized that the robot would imitate and learn from the provided demonstrations. Participants were instructed to provide smooth and natural demonstrations for the robot to learn from. They were also instructed to avoid collisions with the obstacle, and in case of a collision, they were told to continue and attempt to avoid any further collisions. Additionally, participants were asked to keep the liquid as level as possible. If the liquid touched the landmark line, it would be counted as a spill. Once they gripped the bottle, they were not permitted to release it until reaching the placement location.

The participants started with the bottle in the first location (shown in Fig.~\ref{fig:exp2_bottle}), and the robot was set in gravity compensation mode, allowing them to physically guide the robot to pick up the bottle and place it in the designated location. To control the gripper, we used a Wizard-of-Oz technique: participants gave voice commands such as ``open`` or ``close``, and the experimenter executed these commands on the computer to control the gripper. Once the bottle was placed, the data was saved as ROSbag files for offline analysis. The robot was then reset to the initial position, as in Fig.~\ref{fig:exp2_initial}, and the bottle was moved to the second location. This process was repeated until participants provided four demonstrations from each pickup location. The order of the pickup locations (as shown in Fig.~\ref{fig:exp2_top}) was designed to progress from easy to hard; the closer the pickup location is to the robot, the more challenging it becomes to position the robot to pick up the bottle. This phase was referred to as the baseline performance, as it captured participants' demonstrations during their first interaction with the task.

After this baseline phase, all participants underwent a practice period using one of the pickup locations (location 4), chosen for being the most challenging position. Each participant was assigned to one of three training groups as in~\cite{sakr2020training}, with equal practice time allotted for each group. They were encouraged to experiment with different strategies to develop a smooth approach to teaching the robot the task. Following the practice session, a retention test identical to the baseline was conducted to capture participants' performance after practice. For this test, participants provided four demonstrations from each location, picking up the bottle and placing it in the designated spot, to assess their performance post-practice.

    

\subsection{Task Learning}

Several factors contribute to the success rate of a learning policy, such as the number of provided demonstrations, their distribution over the task space, the learning algorithm, and demonstration quality. Here, we focus on the quality of the provided demonstrations while keeping all other factors fixed (i.e., the number of demonstrations, the distribution, and the learning model). 
For the learning algorithm, we used a Task Parameterized Gaussian Mixture Model (TP-GMM) combined with Gaussian Mixture Regression (GMR) for task learning. TP-GMMs have been extensively used in the LfD literature~\cite{pervez2018learning, osa2018algorithmic}, providing good generalization with a limited set of demonstrations.

TP-GMM models a task using task parameters defined by a sequence of coordinate frames, each represented by a matrix for orientation and a vector for origin relative to a global frame~\cite{calinon2016tutorial}. A Gaussian Mixture Model (GMM) is fitted to the data in each local frame, with the parameters estimated using an expectation-maximization algorithm. To generate a trajectory, the local models are transformed back into the global frame and combined into a global model through a product of Gaussians. The resulting trajectory is then used to derive a feasible joint space trajectory. To ensure smooth and accurate task execution, both the singularity-robust inverse~\cite{nakamura1986inverse} and a closed-loop inverse kinematics (CLIK) algorithm~\cite{colome2014closed} are employed. Given the redundancy in a 7-DOF manipulator, the null space is utilized to control joint space motion without affecting task-space motion, using the closest human demonstration as a policy~\cite{liegeois1977automatic}. This comprehensive approach allows for accurate task reproduction while avoiding kinematic singularities and minimizing errors.

\subsection{Evaluation of Learning and Generalization Performance}


\subsubsection{Experiment 1: Button-pressing task evaluation}
Model performance was evaluated on the same set of demonstrated tasks (task performance), as in Fig.~\ref{fig:exp1_eval_task}, and on new target positions (generalized performance), as in Fig.~\ref{fig:exp1_eval_general}. Generalization was evaluated by discretizing the face of the box in Fig.~\ref{fig:exp_initial} into a 7x7 grid, resulting in a total of 49 new target positions. The grid size was determined based on the dimensions of the PR2 arm's tip and the box to avoid overlap between targets. The arm tip's dimensions are $(W = 2.1 \text{ cm}, L = 2.2 \text{ cm}, H = 3.5 \text{ cm})$, and the box is a cube with edges of 26 cm. We specified the target point as a sphere with a 3 cm diameter centered on the button.

The learned trajectory was considered successful if it reached the goal position (within the goal sphere) while avoiding self-collisions and collisions with the box. To account for the non-zero size of the end-effector tip, we considered the robot to have reached the goal if any point of the tip touched the goal sphere. The success rate for each trial was calculated by dividing the number of successful trajectories by the total number of tested points (nine in the task performance test and 49 in the generalization test).

\subsubsection{Experiment 2: Pick-and-place task evaluation}
\begin{figure}[t]
    \centering
    \begin{subfigure}[t]{0.48\textwidth} 
        \centering
        \begin{subfigure}[t]{0.48\textwidth}
            \centering
            \includegraphics[width=\textwidth]{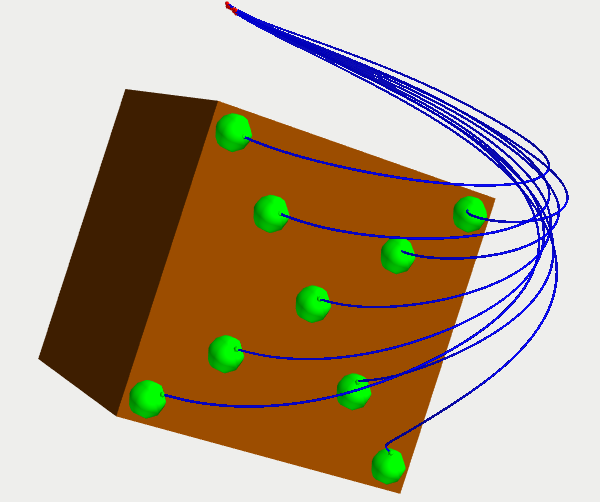}
            \caption{(a) Same task states}
            \label{fig:exp1_eval_task}
        \end{subfigure}
        \hfill
        \begin{subfigure}[t]{0.48\textwidth}
            \centering
            \includegraphics[width=\textwidth]{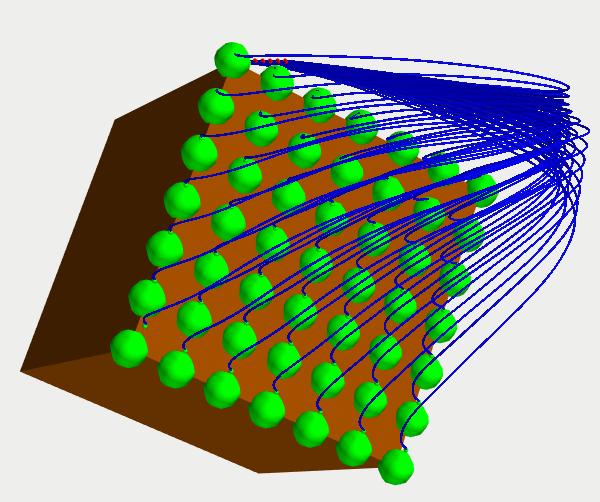}
            \caption{(b) Generalized states}
             \label{fig:exp1_eval_general}
        \end{subfigure}
        \caption{(I) First Experiment (button pressing task with PR2)}
        \label{fig:exp1_evaluation}
    \end{subfigure}
    \hfill
    \begin{subfigure}[t]{0.47\textwidth} 
        \centering
        \begin{subfigure}[t]{0.47\textwidth}
            \centering
            \includegraphics[width=\textwidth]{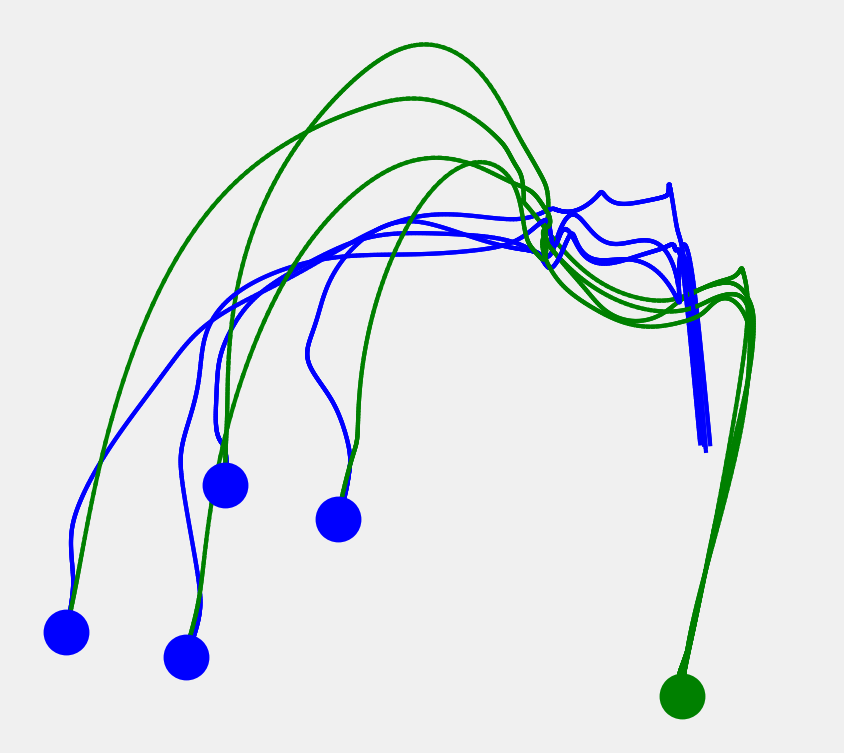}
            \caption{(d) Same task states}
             \label{fig:exp2_eval_task}
        \end{subfigure}
        \hfill
        \begin{subfigure}[t]{0.47\textwidth}
            \centering
            \includegraphics[width=\textwidth]{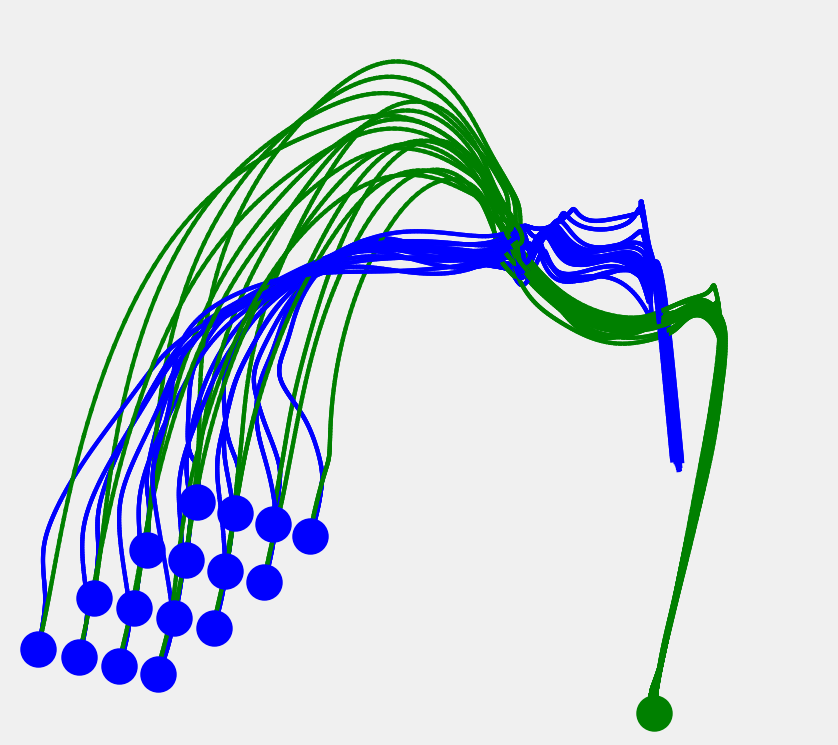}
            \caption{(e) Generalized states}
            \label{fig:exp2_eval_general}
        \end{subfigure}
        \caption{(II) Second Experiment (pick and place task with UR5)}
        \label{fig:exp2_evaluation}
    \end{subfigure}
    \caption{The tasks simulation includes both the same task states and the generalized states used to evaluate model performance. (I) In the button-pressing task, the buttons are represented as green spheres. (II) In the pick-and-place task, the pickup and place locations are also represented by spheres. The blue segment of the trajectory represents the movement from the initial position to the pickup location, while the green segment represents the movement from the pickup to the place location.}
    \label{fig:exp_evaluation}
\end{figure}

Model performance was evaluated by generating trajectories to pick and place the bottle from the same demonstrated locations (task performance), as in Fig.~\ref{fig:exp2_eval_task}, and from new pickup locations (generalized performance), as in Fig.~\ref{fig:exp2_eval_general}. The generalized pickup locations was created by discretizing the pickup space into a 4x4 grid, resulting in 16 new pickup locations. The grid size was chosen based on the UR5 robot's reachability space and the diameter of the experimental bottle (7 cm).

The learned trajectory was considered successful if the bottle was picked up and placed without ``spillage``. Spillage was detected by monitoring deviations in upright orientation, acceleration, and jerk, which serve as proxies for potential liquid movement. We normalized the orientation deviation, jerk, and acceleration values to assess spillage risk. As the jerk, acceleration, and orientation deviation values approach zero, the success score approaches 1, provided the bottle was picked up and placed correctly. However, as jerk, acceleration, and/or orientation deviation increase, the success score is penalized accordingly. The Overall Success was computed using a weighted sum of the pick-and-place success and the normalized spillage proxies, as shown in Equation~\ref{eq:exp2_success}.

\begin{equation}
\text{Overall Success} = 
\begin{cases}
0, & \text{if the bottle was not picked up} \\[10pt]
0.5 \times (S_p + S_l) + 0.5 \times \frac{(1 - J_n) + (1 - A_n) + (1 - O_n)}{3}, & \text{if the bottle was picked up successfully}
\end{cases}
\label{eq:exp2_success}
\end{equation}

\noindent where \( S_p \) denotes the pickup success, with \( S_p = 0.5 \) if the bottle was successfully picked up and \( S_p = 0 \) otherwise. \( S_l \) represents the place success, with \( S_l = 0.5 \) if the bottle was placed correctly and \( S_l = 0 \) otherwise. The values \( J_n \), \( A_n \), and \( O_n \) correspond to the normalized metrics for jerk, acceleration, and orientation deviation, respectively, 

The Phase Success score was then calculated by averaging the overall success rates across all tested locations, as defined in Equation~\ref{eq:exp2_phase_success}. In the task performance test, this phase success was calculated across four locations, while in the generalization test, it was calculated across 16 locations.

\begin{equation}
\text{Phase\_Success} = \frac{\sum_{i=1}^{N} \text{Overall\_Success}_i}{N}
\label{eq:exp2_phase_success}
\end{equation}

\noindent where Phase\_Success represents the average success score across all tested locations within the phase. 
Overall\_Success$_i$is the individual success score for each location \( i \), calculated as in Equation~\ref{eq:exp2_success}. \( N \) denotes the total number of tested locations.


%% file: Text/hypotheses.tex
\begin{figure*}[t]
    \centering
    \begin{subfigure}[t]{0.9\textwidth}
        \centering
        \includegraphics[width=\textwidth]{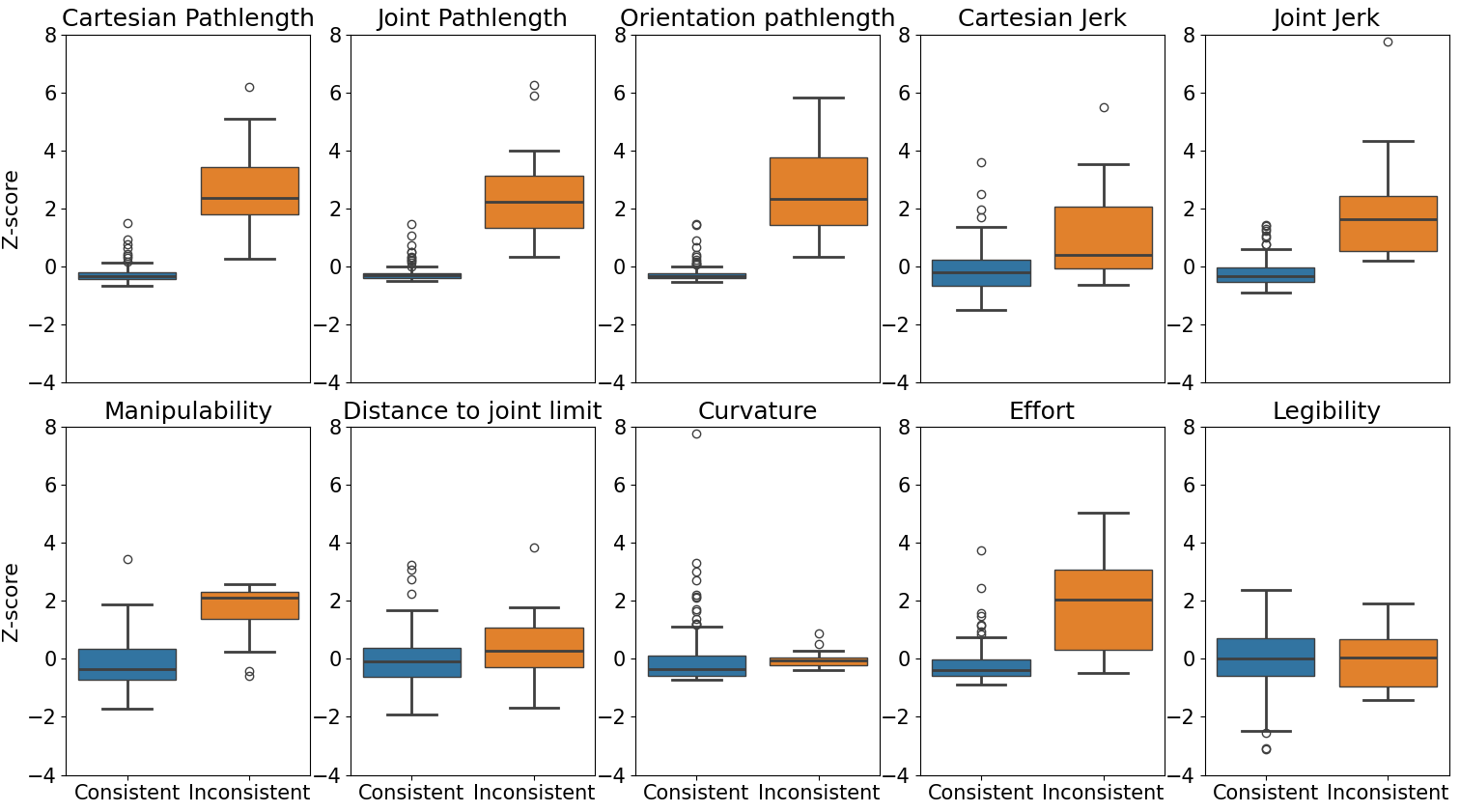}
        \caption{(a) First experiment (button-pressing task with PR2)}
        \label{fig:exp1_features}
    \end{subfigure}
    \hspace{3cm}
    \begin{subfigure}[t]{0.9\textwidth}
        \centering
        \includegraphics[width=\textwidth]{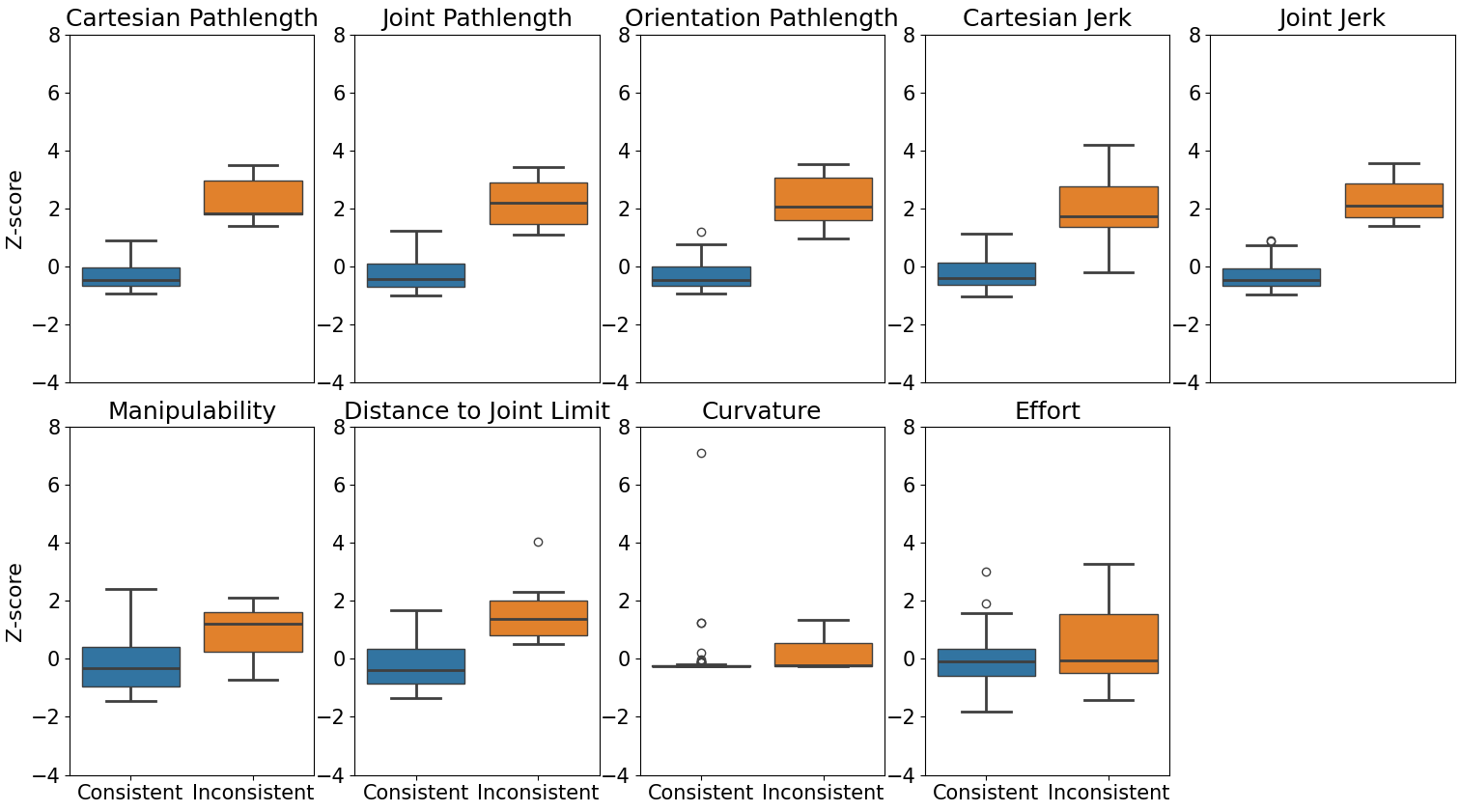}
        \caption{(b) Second experiment (pick-and-place task with UR5)}
        \label{fig:exp2_features}
    \end{subfigure}
    
    \caption{The standardized feature values in the resulting two clusters of demonstrations. The first cluster (blue) includes consistent demonstrations and the second cluster (orange) includes inconsistent demonstrations.}
\label{fig:clusters_features}
\end{figure*}

We expect that the demonstrations' consistency in terms of the proposed measures is a key identifier of the quality, and consequently, of task performance and generalized performance. Moreover, we believe that practice is crucial for improving the quality of demonstrations, which in turn enhances task performance and generalized performance. Based on these expectations, we formulate the following hypotheses:

\begin{itemize}
\item \textbf{H1-a}: The consistency of demonstrations in terms of the proposed metrics significantly contributes to the task performance success rate.
\item \textbf{H1-b}: The consistency of demonstrations in terms of the proposed metrics significantly contributes to the generalized performance success rate.
\item \textbf{H2}: Participants' practice significantly improves the consistency of demonstrations.
\item \textbf{H3-a}: Participants' practice significantly improves the task performance success rate.
\item \textbf{H3-b}: Participants' practice significantly improves the generalized performance success rate.
\end{itemize}



%% file: Text/results.tex
To evaluate our hypotheses, we assessed the consistency of the demonstrations provided by each participant. We employed a K-means clustering algorithm~\cite{hartigan1979k} to categorize each set of demonstrations into one of two clusters: a \textit{consistent} cluster and an \textit{inconsistent} cluster, as shown in Fig.~\ref{fig:overview}. For each trial in the button-pressing experiment, the range of the ten metrics listed in Table~\ref{tab:metrics} was calculated and used as input for the clustering algorithm (multi-dimensional clustering). The rationale for using the range is that consistent demonstrations are expected to exhibit smaller variations in metric values, while inconsistent demonstrations are expected to show larger variations.

Fig.~\ref{fig:exp1_features} shows the standardized metric values for both the consistent and inconsistent clusters in the button-pressing task. The inconsistent cluster displayed a larger range across most metrics, while the consistent cluster exhibited a smaller range. Specifically, the inconsistent cluster had a broader distribution in all pathlength, jerk, and effort metrics. In contrast, the distribution for distance to joint limits, curvature, and legibility was similar between the two clusters. Additionally, the inconsistent cluster showed higher median values for pathlength, jerk, manipulability, and effort metrics, indicating that demonstrations in the inconsistent trials involved more variability, longer paths, greater jerk, more singular poses, and higher energy expenditure. Conversely, the consistent trials, with lower median values for these metrics, demonstrated smoother, more conservative, and less energy-intensive movements.

In the pick-and-place experiment, we applied the same approach by using the range of the proposed metrics from each participant’s baseline and retention phases as input features for the K-means clustering algorithm. Unlike the button-pressing experiment, legibility was excluded here since there was only one goal (placement) location. Fig.~\ref{fig:exp2_features} shows the standardized metric values for both clusters. Similar to the button-pressing experiment, the inconsistent cluster had a higher range for most metrics compared to the consistent cluster, which had a smaller range. Moreover, the distribution of the range for pathlength, jerk, curvature, and effort metrics differed significantly between the two clusters, while the distribution for manipulability and distance to joint limits was more similar between them.

\begin{figure*}[t]
    \centering
    \begin{subfigure}[t]{0.48\textwidth}
        \centering
        \includegraphics[width=\textwidth]{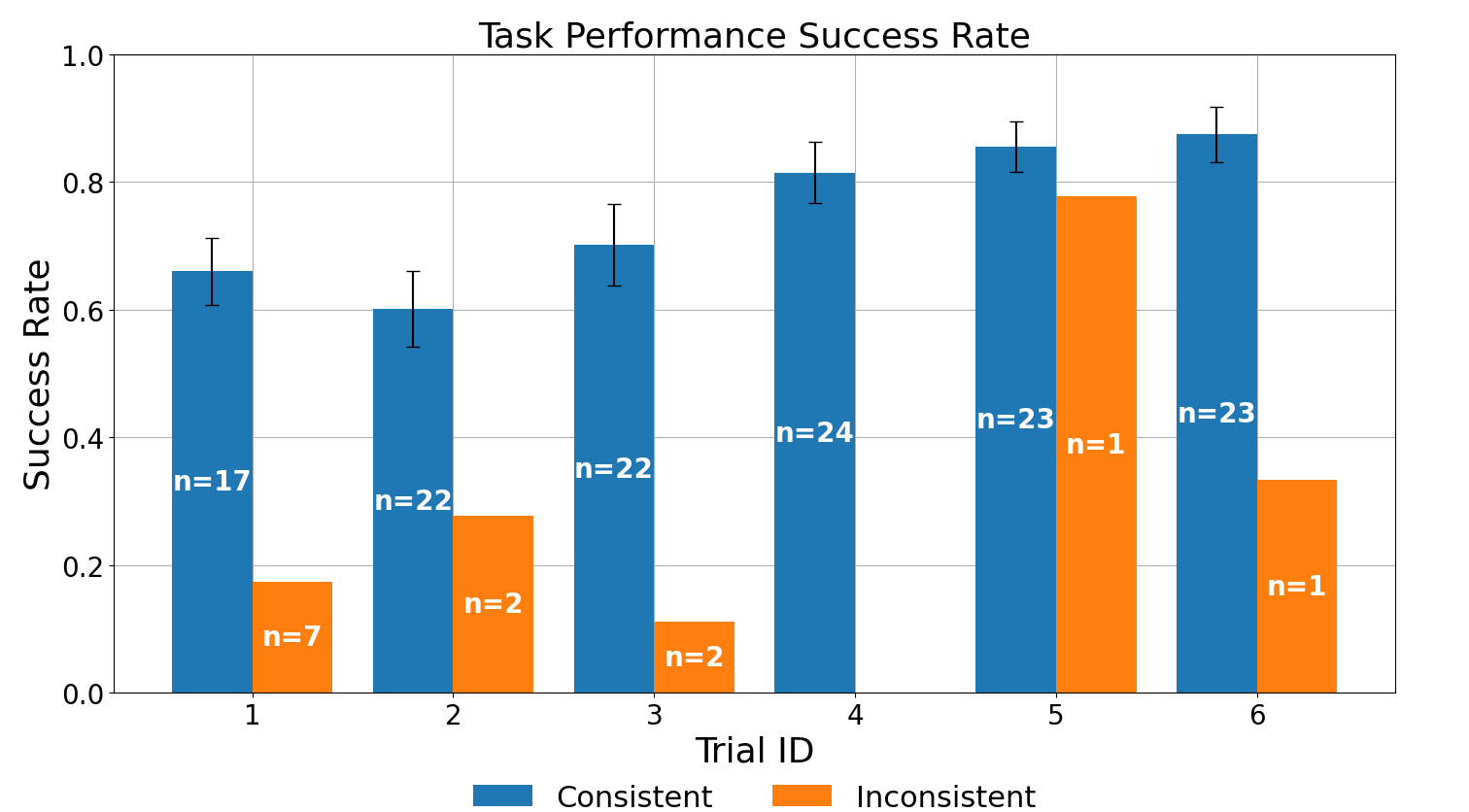}
        \caption{(a) First experiment (button-pressing task with PR2)}
        \label{fig:exp1_task_success}
    \end{subfigure}
    \hfill
    \begin{subfigure}[t]{0.48\textwidth}
        \centering
        \includegraphics[width=\textwidth]{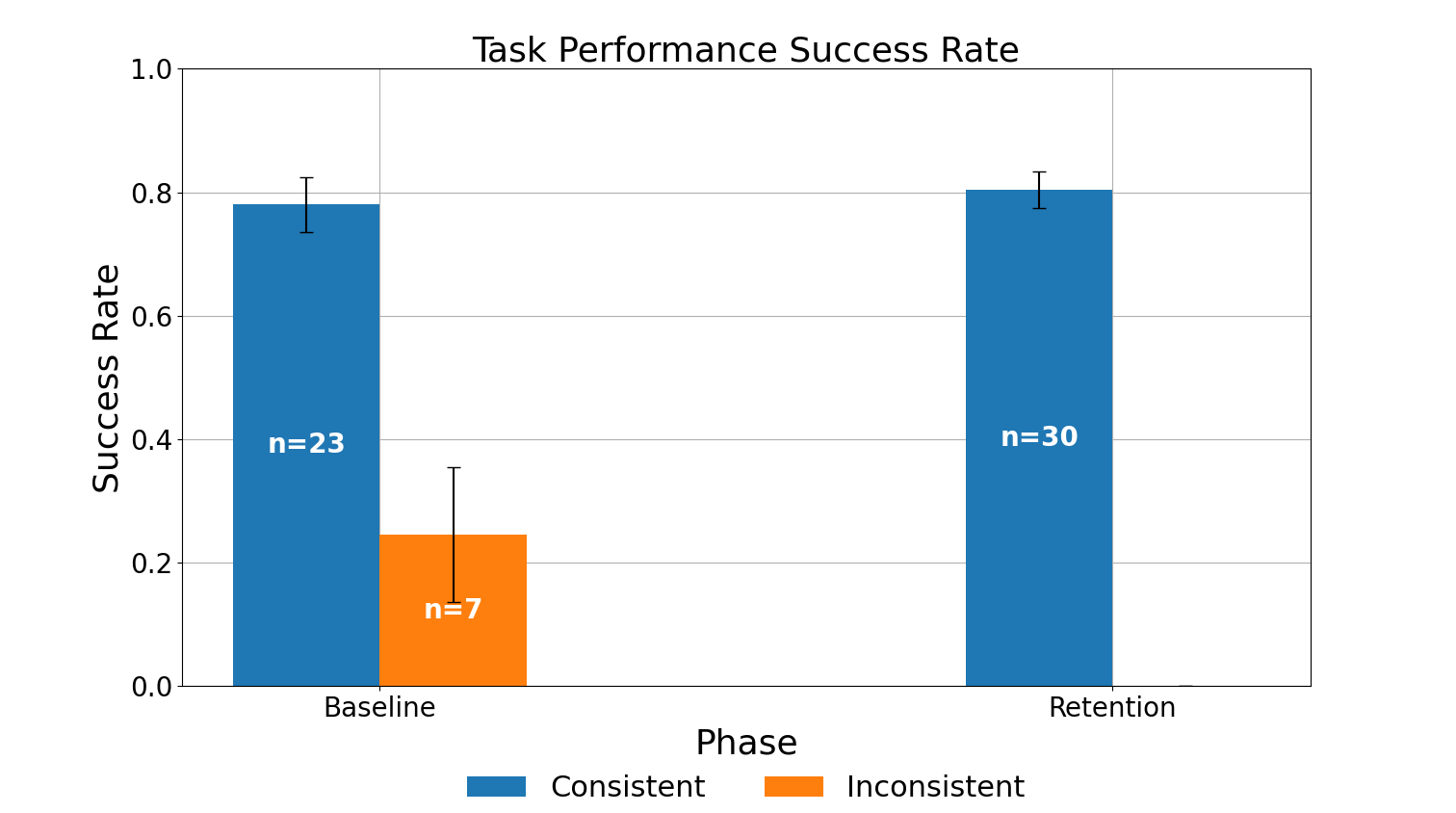}
        \caption{(b) Second experiment (pick and place task with UR5)}
        \label{fig:exp2_task_success}
    \end{subfigure}
    
    \caption{Average task performance success rate across trials/phases for both consistent and inconsistent groups.}
\label{fig:task_success}
\end{figure*}

\subsection{Consistency impact on the task and generalized success rate}
After categorizing the consistent and inconsistent clusters, we investigated the task performance success rate for both clusters. As shown in Fig.\ref{fig:task_success}, the task performance success rate when training the model with consistent trials is higher than when training with inconsistent ones. For the button-pressing experiment results, a one-way ANOVA was performed to investigate the impact of consistency on the success rate. Since the number of inconsistent sets in each trial is low, we aggregated the inconsistent and consistent sets across all trials. This analysis yields the relationship between the two vectors: one for the task success rate and the other for the consistency label (0 or 1). We found that the consistent group had a significantly higher success rate than the inconsistent group $(F(1,142) = 45.46, p < 0.001)$, with an average success rate of $(0.76 \pm 0.02)$\footnote{$(mean \pm SE)$} in the consistent group and $(0.24 \pm 0.08)$ in the inconsistent group. Similarly, a one-way ANOVA was performed on the data from the pick-and-place experiment. We found that the consistent group had a significantly higher success rate than the inconsistent group $(F(1,58) = 44.01, p < 0.001)$ with an average success rate of $(0.79 \pm 0.03)$ in the consistent group and $(0.25 \pm 0.11)$ in the inconsistent group. These results support \textbf{H1-a}.

\begin{figure*}[t]
    \centering
    \begin{subfigure}[t]{0.48\textwidth}
        \centering
        \includegraphics[width=\textwidth]{Images/task_performance_success_rate_py.png}
        \caption{(a) First experiment (button-pressing task with PR2)}
        \label{fig:exp1_generalized_success}
    \end{subfigure}
    \hfill
    \begin{subfigure}[t]{0.48\textwidth}
        \centering
        \includegraphics[width=\textwidth]{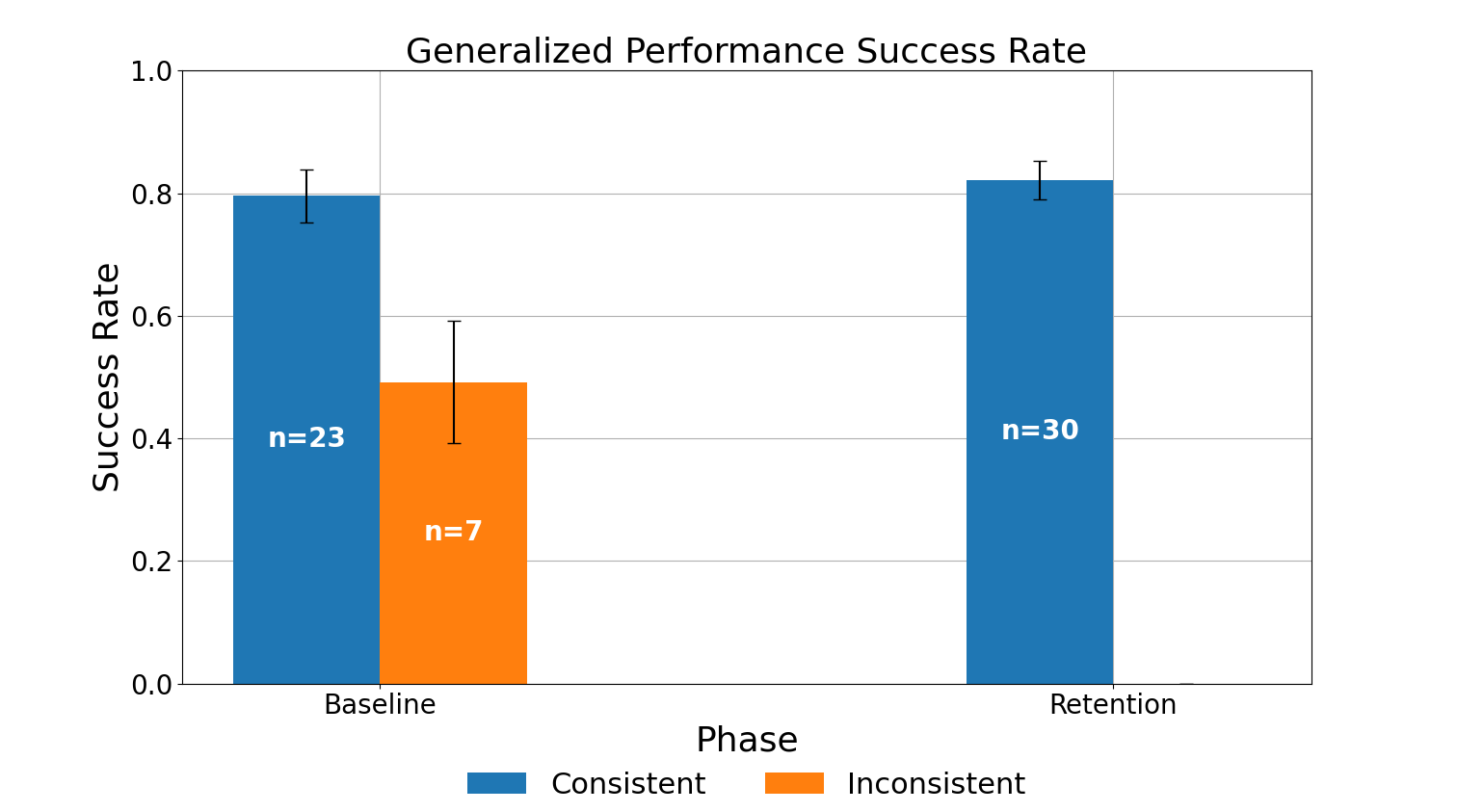}
        \caption{(b) Second experiment(pick and place task with UR5)}
        \label{fig:exp2_generalized_success}
    \end{subfigure}
    
    \caption{Average generalized performance success rate across trials/phases for both consistent and inconsistent groups.}
\label{fig:generalized_success}
\end{figure*}

We conducted the same ANOVA analysis on the generalized performance success rate and found a similar result, as shown in Fig.\ref{fig:generalized_success}. In the button-pressing experiment, the consistent group had a significantly higher success rate than the inconsistent group $(F(1,143) = 36.57, p < 0.001)$, with an average success rate of $(0.63 \pm 0.02)$ in the consistent group and $(0.22 \pm 0.06)$ in the inconsistent group. Similarly, in the pick-and-place experiment, the consistent group had a significantly higher success rate than the inconsistent group $(F(1,58) = 15.50, p < 0.001)$, with an average success rate of $(0.81 \pm 0.03)$ in the consistent group and $(0.49 \pm 0.10)$ in the inconsistent group, supporting \textbf{H1-b}.

\subsection{Impact of practice on demonstration consistency}
To assess \textbf{H2} and investigate the impact of practice on the consistency of the demonstrations, a MANOVA (Multivariate Analysis of Variance) was performed to investigate the impact of practice on the range of the ten metrics, as detailed in Table~\ref{tab:manova_univariate_results}. In the button-pressing experiment, the multivariate tests revealed a significant effect of practice (trials) on the consistency of the demonstrations $(Roy's Largest Root = 0.559, F(10, 110) = 6.144, p < 0.001, \eta_p^2 = 0.358)$, with 35.8\% of the variance in the range of metrics explained by the effect of the trial. Moreover, Fig.~\ref{fig:exp1_task_success} shows that the number of inconsistent trials that are also first trials is higher compared to later trials, indicating that consistency improves with practice. Similarly, we conducted MANOVA on the range of nine metrics of the pick-and-place experiment data to investigate the practice/training impact on the consistency of these metrics. The test revealed a significant effect of practice/training on the consistency of the demonstrations $(Roy's Largest Root = 0.559, F(9, 21) = 4.35, p < 0.005, \eta_p^2 = 0.651)$, with 65.1\% of the variance in the range of metrics explained by the effect of practice, supporting \textbf{H2}.

Univariate tests on the button-pressing experiment showed significant effects of the trials on the following range of metrics: jerk in Cartesian space, path orientation length, and effort $(p < 0.05)$. They also revealed marginally significant effects of the trials on the range of metrics for pathlength in Cartesian space, pathlength in joint space, and manipulability $(p = 0.07)$. Post hoc pairwise comparisons revealed that participants may need more time to achieve significantly better values for some metrics than others. For example, the participants showed a significant improvement after one trial in terms of pathlength in Cartesian and joint space, manipulability, and curvature. For metrics like jerk in Cartesian and joint space and effort, two trials were required for participants to better demonstrate the task and achieve significantly lower jerk values. Additionally, for legibility, only the first and last trials were significantly different, indicating that participants may need more time to generate legible demonstrations. Similarly in the pick-and-place experiment, univariate tests showed a significant effect of the practice on the range of the following metrics: pathlength in Cartesian and joint space, orientation pathlength and jerk in Cartesian and joint space $p < 0.01$. 

\begin{table*}[ht]
\centering
\caption{Results of MANOVA and Univariate Tests for Button-Pressing and Pick-and-Place Experiments}
\label{tab:manova_univariate_results}
\begin{tabular}{lcccc}
\toprule
\textbf{Experiment} & \textbf{Roy's Largest Root} & \textbf{F-value} & \textbf{p-value} & \textbf{Partial Eta Squared ($\eta_p^2$)} \\
\midrule
\multicolumn{5}{c}{\textbf{MANOVA Results}} \\
\midrule
Button-Pressing   & 0.559  & 6.144  & < 0.001 & 0.358 \\
Pick-and-Place    & 0.559  & 4.350  &  0.003 & 0.651 \\
\midrule
\multicolumn{5}{c}{\textbf{Univariate Test Results}} \\
\midrule
\textbf{Metric} & \textbf{Experiment} & \textbf{F-value} & \textbf{p-value} & \textbf{Partial Eta Squared ($\eta_p^2$)} \\
\midrule
Path Orientation Length       & Button-Pressing   & 2.40   & 0.04   & 0.10 \\
Jerk in Cartesian Space       & Button-Pressing   & 4.12   & 0.002  & 0.15 \\
Effort                        & Button-Pressing   & 3.07   & 0.012  & 0.12 \\
Pathlength in Cartesian Space & Button-Pressing   & 2.10   & 0.07   & 0.08 \\
Pathlength in Joint Space     & Button-Pressing   & 2.10   & 0.07   & 0.08 \\
Manipulability                & Button-Pressing   & 2.12   & 0.07   & 0.08 \\
Pathlength in Cartesian Space & Pick-and-Place    & 13.23  & 0.001  & 0.31 \\
Pathlength in Joint Space     & Pick-and-Place    & 7.90   & 0.009  & 0.21 \\
Path Orientation Length       & Pick-and-Place    & 9.52   & 0.004  & 0.25 \\
Jerk in Cartesian Space       & Pick-and-Place    & 12.51  & 0.001  & 0.30 \\
Jerk in Joint Space           & Pick-and-Place    & 7.87   & 0.009  & 0.21 \\
\bottomrule
\end{tabular}
\end{table*}

\subsection{Practice impact on the task and generalized success rate}
We hypothesized that practice improves the overall success rate for both task performance and generalized performance. This is because the consistency (as shown earlier) and other factors improve over time. To evaluate this hypothesis (\textbf{H3-a} and \textbf{H3-b}), a repeated ANOVA with a Greenhouse-Geisser correction was performed to account for the violation of the sphericity assumption, exploring the impact of practice (trials) on the success rate of task performance, as shown in Table~\ref{tab:task_performance_anova}. We found that practice has a significant impact on the success rate $(F(1,23) = 23.47, p<0.001, \eta_p^2 = 0.51)$. The success rate improved from $(0.52 \pm 0.07)$ in the first trial to $(0.85 \pm 0.05)$ in the last trial. In addition, the post hoc pairwise comparison of the trials shows that trials 1, 2, and 3 have significantly lower success rates than trials 4, 5, and 6, while there is no significant difference within the first three trials or within the last three trials. This highlights that the significant improvement is revealed in the second session on later days, which agrees with Walker et al.~\cite{walker2003sleep}. The same analysis was applied to the pick-and-place experiment task success rate. The repeated ANOVA showed task success rate in retention (after training) is significantly higher than the success rate in the baseline (before training) $(F(1,29) = 9.09, p<0.01, \eta_p^2 = 0.24)$ with an average success rate of $(0.66 \pm 0.06)$ in the baseline and $(0.80 \pm 0.03)$ in the retention, supporting \textbf{H3-a}.

\begin{table}[ht]
\centering
\caption{Repeated ANOVA Results for Task Performance Success Rates}
\label{tab:task_performance_anova}
\begin{tabular}{lcccc}
\toprule
\textbf{Experiment} & \textbf{Phases/Trials} & \textbf{F-value} & \textbf{p-value} & \textbf{Partial Eta Squared ($\eta_p^2$)} \\
\midrule
Button-Pressing   & 6 Trials & 23.47  & $< 0.001$ & 0.51 \\
Pick-and-Place    & 2 Phases & 9.09   & $< 0.01$  & 0.24 \\
\bottomrule
\end{tabular}

\vspace{0.5em}

\begin{tabular}{lcc}
\toprule
\textbf{Trial/Phase} & \textbf{Button-Pressing} & \textbf{Pick-and-Place} \\
\midrule
Initial (First Trial/Baseline) & $0.52 \pm 0.07$ & $0.66 \pm 0.06$ \\
Final (Last Trial/Retention)   & $0.85 \pm 0.05$ & $0.80 \pm 0.03$ \\
\bottomrule
\end{tabular}


\end{table}

A repeated ANOVA was conducted on the generalized success rate (as detailed in Table~\ref{tab:generalized_performance_anova}) and we found a similar result. In the button-pressing experiment, practice has a significant impact on the generalized success rate $(F(1,23) = 6.16, p<0.05, \eta_p^2 = 0.21)$. The success rate improved from $(0.51 \pm 0.06)$ in the first trial to $(0.70 \pm 0.03)$ in the last trial. In the pick-and-place experiment, training has a significant impact on the generalized success rate $(F(1,29) = 5.87, p<0.05, \eta_p^2 = 0.17)$. The success rate improved from $(0.73 \pm 0.05)$ in the baseline(before training) to $(0.82 \pm 0.03)$ in the retention(after training), supporting \textbf{H3-b}.

The post hoc pairwise comparison of the trials in the button-pressing experiment shows that trials 1 and 2 are significantly less successful than trial 6, while there is no significant difference among the intermediate trials. This indicates that improvement in the generalized success rate is slower than in the task performance success rate, and more time is needed for the participants to achieve a good quality of demonstrations that attain a better generalized success rate.

\begin{table}[ht]
\centering
\caption{Repeated ANOVA Results for Generalized Success Rates}
\label{tab:generalized_performance_anova}
\begin{tabular}{lcccc}
\toprule
\textbf{Experiment} & \textbf{Phases/Trials} & \textbf{F-value} & \textbf{p-value} & \textbf{Partial Eta Squared ($\eta_p^2$)} \\
\midrule
Button-Pressing   & 6 Trials & 6.16  & $< 0.05$ & 0.21 \\
Pick-and-Place    & 2 Phases & 5.87   & $< 0.05$  & 0.17 \\
\bottomrule
\end{tabular}

\vspace{0.5em}

\begin{tabular}{lcc}
\toprule
\textbf{Trial/Phase} & \textbf{Button-Pressing} & \textbf{Pick-and-Place} \\
\midrule
Initial (First Trial/Baseline) & $0.51 \pm 0.06$ & $0.73 \pm 0.05$ \\
Final (Last Trial/Retention)   & $0.70 \pm 0.03$ & $0.82 \pm 0.03$ \\
\bottomrule
\end{tabular}


\end{table}

\subsection{The importance of each metric to the task success rate}
Thus far, we have demonstrated that consistency is critical for achieving higher success rates and that practice improves demonstration quality, which, in turn, enhances success rates. We have also observed that the proposed metrics show different patterns across trials and clusters. In this section, we explore the importance of each metric in determining success rates by investigating the question of \textit{What are the key aspects to evaluate in demonstrations before learning?}

\begin{figure*}[t]
    \centering
    \begin{subfigure}[t]{0.48\textwidth}
        \centering
        \includegraphics[width=\textwidth]{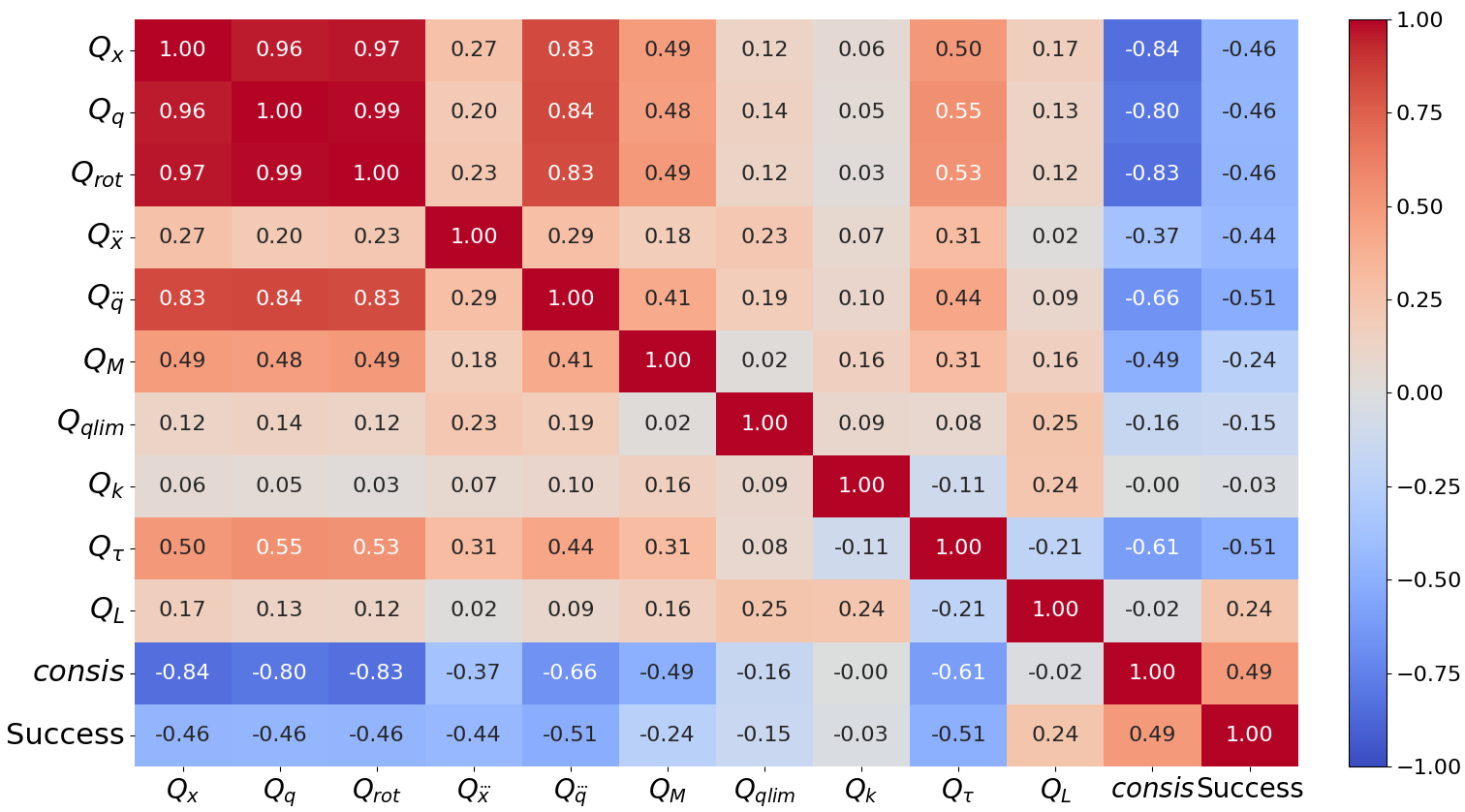}
        \caption{(a) First experiment (button-pressing task with PR2)}
        \label{fig:exp1_corr_heatmap}
    \end{subfigure}
    \hfill
    \begin{subfigure}[t]{0.48\textwidth}
        \centering
        \includegraphics[width=\textwidth]{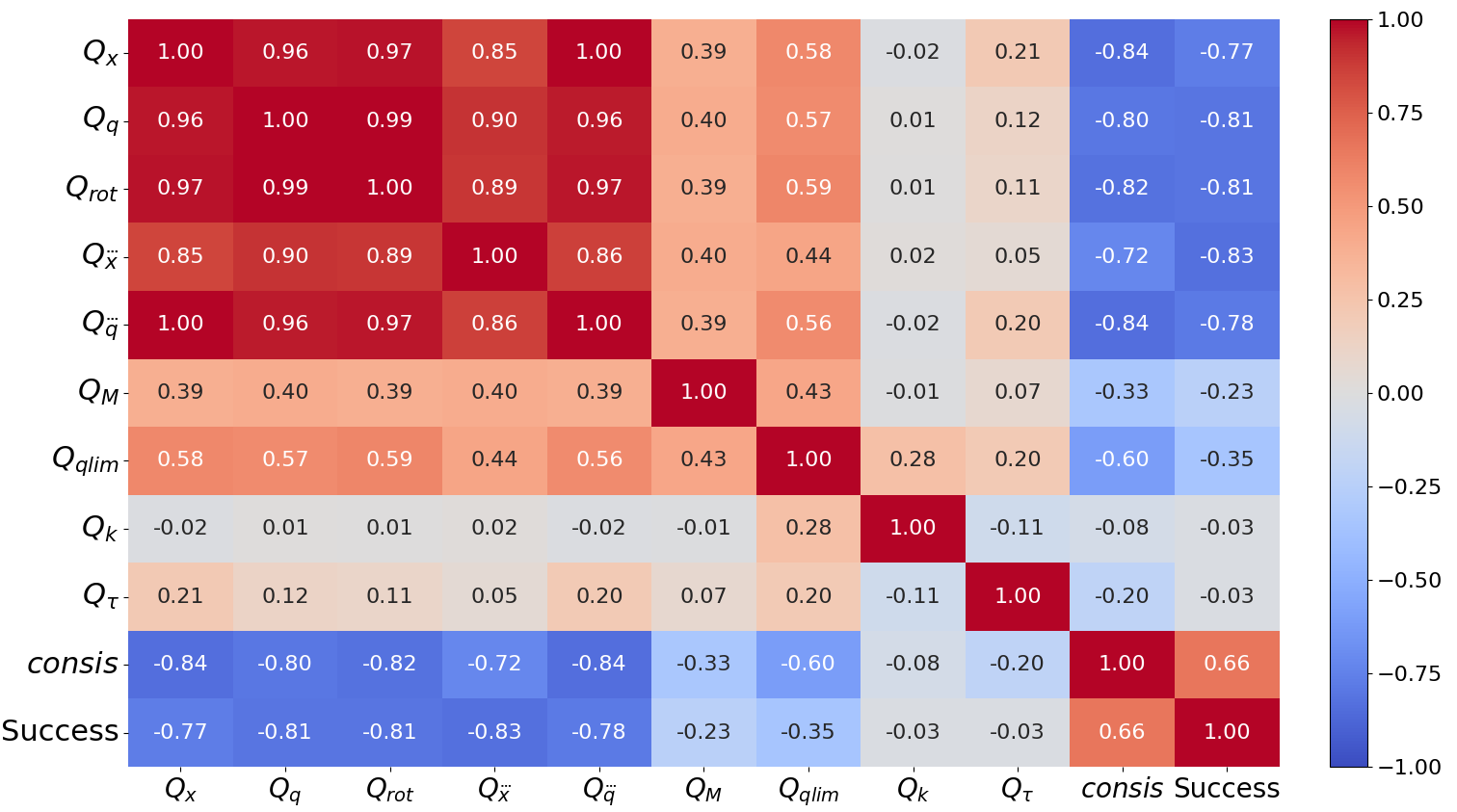}
        \caption{(b) Second experiment (pick and place task with UR5)}
        \label{fig:exp2_corr_heatmap}
    \end{subfigure}
    
    \caption{Heatmap shows the correlation matrix of the task success rate, and the range of the metrics and the consistency. The metrics are defined in Table~\ref{tab:metrics}}
\label{fig:corr_heatmap}
\end{figure*}

To address this, we conducted a correlation analysis between the task success rate and the range of ten proposed metrics. Additionally, we included a binary parameter, consistency, to examine its contribution to the success rate. Fig.~\ref{fig:corr_heatmap} illustrates a heatmap showing the correlations between the metrics and the success rate in both experiments. In both experiments, pathlength, jerk, and consistency exhibit the highest correlations with task success rate. Specifically, pathlength and jerk have negative correlations with success rate, indicating that greater variability reduces success, while consistency shows a positive correlation, suggesting that more consistent demonstrations improve performance.

Other terms vary between experiments. Effort has a high correlation with success rate in the button-pressing experiment but a lower correlation in the pick-and-place. Conversely, the distance to joint limits correlates more strongly with success rate in the pick-and-place experiment. We also found that some metrics, like pathlength, are highly correlated with each other, and consistency correlates strongly with pathlength, jerk in joint space, and effort. This pattern aligns with Fig.~\ref{fig:clusters_features}, where pathlength, jerk, and effort show the most distinct patterns between clusters.

Motivated by these correlations, we performed multiple regression analysis, including interaction terms, to examine how changes in correlated metrics affect the success rate. The range of the proposed metrics collectively predicts 70.2\% and 88.6\% of the task success rate in the two experiments, respectively. Notably, the interactions among the ranges of pathlength and jerk metrics are the most significant factors with high coefficients for estimating the success rate in both experiments. Details of the regression models and the analysis are provided in Appendix~\ref{appendix:reg_task}.

\subsection{The importance of each metric to the generalized success rate}

\begin{figure*}[t]
    \centering
    \begin{subfigure}[t]{0.48\textwidth}
        \centering
        \includegraphics[width=\textwidth]{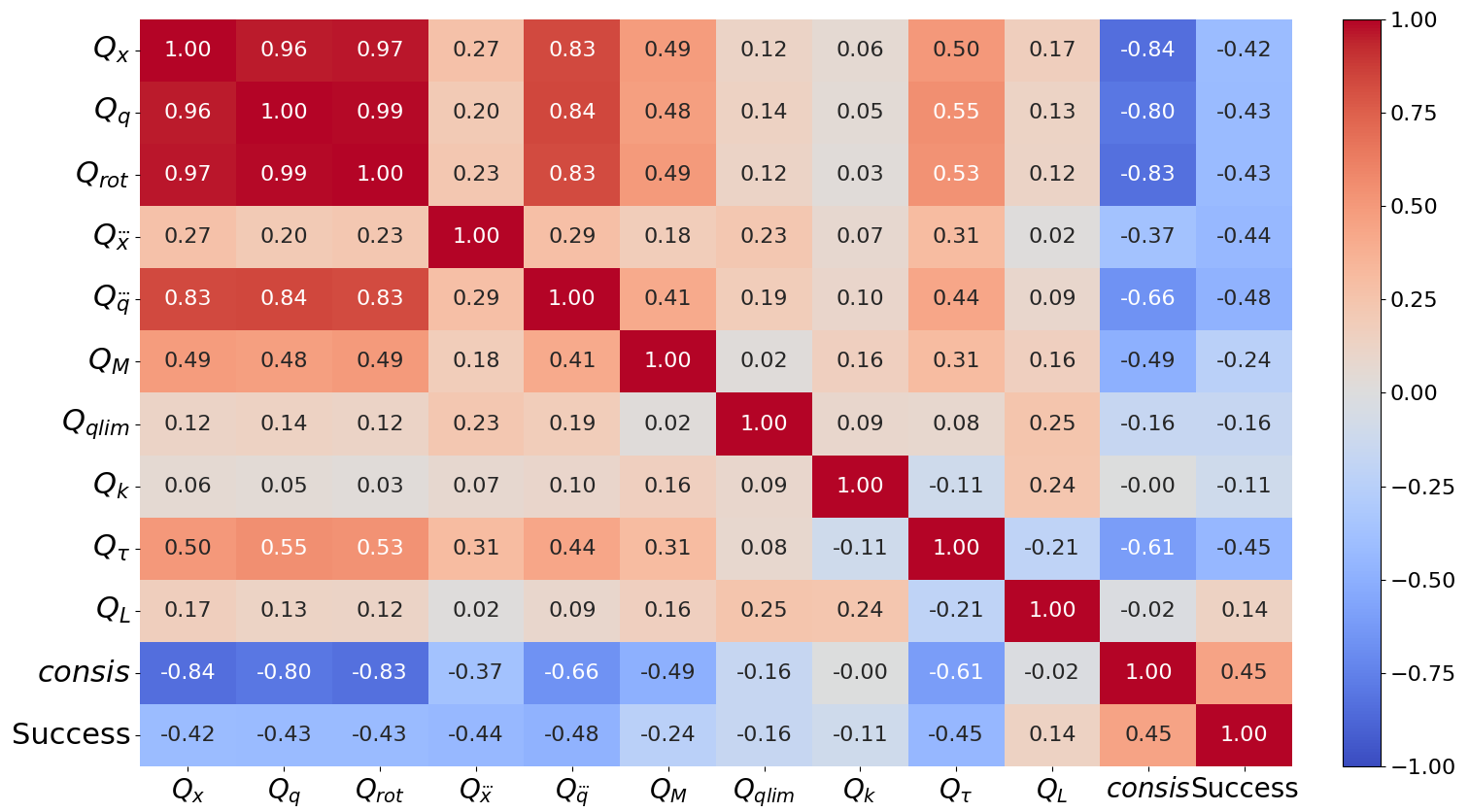}
        \caption{(a) First experiment (button-pressing task with PR2)}
        \label{fig:exp1_corr_heatmap}
    \end{subfigure}
    \hfill
    \begin{subfigure}[t]{0.48\textwidth}
        \centering
        \includegraphics[width=\textwidth]{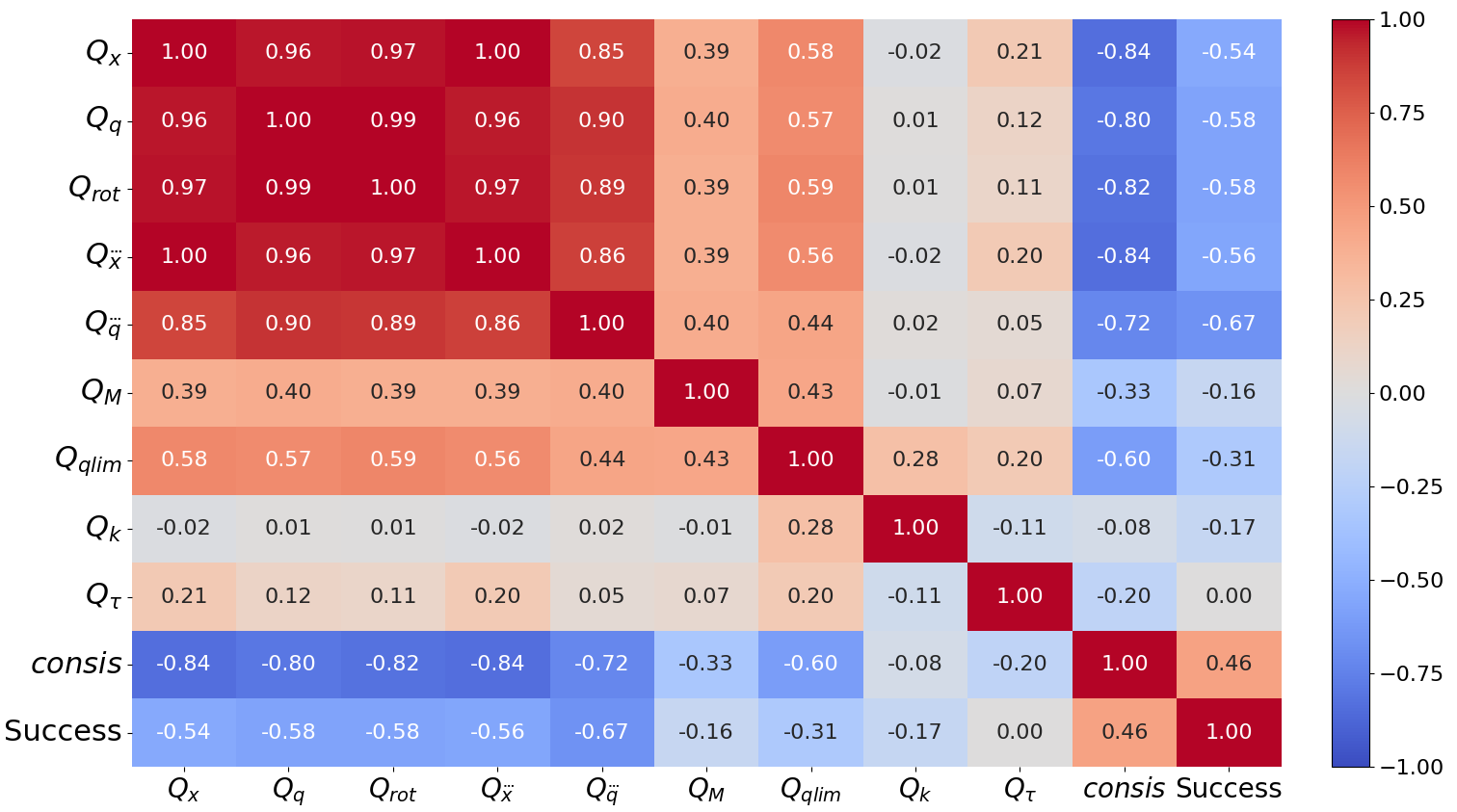}
        \caption{(b) Second experiment (pick and place task with UR5)}
        \label{fig:exp2_corr_heatmap}
    \end{subfigure}
    
    \caption{Heatmap shows the correlation matrix of the generalized success rate, and the range of the metrics and the consistency. The metrics are defined in Table~\ref{tab:metrics}}
\label{fig:gen_corr_heatmap}
\end{figure*}

In this section, we examine the relationship between the generalized success rate and demonstration consistency using the proposed measures. Similar to the analysis of the task success rate, we first conducted a correlation analysis. Fig.~\ref{fig:gen_corr_heatmap} presents a heatmap showing the correlation between the generalized success rate, the range of the proposed measures, and the binary consistency variable across the two experiments. Pathlength, jerk, and consistency measures exhibit the highest correlation with the generalized success rate, mirroring the pattern observed for task success. Similarly, effort shows a high correlation with the generalized success rate in the button-pressing experiment but a weaker correlation in the pick-and-place experiment. Conversely, distance to joint limits has a stronger correlation with the generalized success rate in the pick-and-place experiment than in the button-pressing. Several measures are also strongly correlated with one another, which motivated the use of regression analysis with interaction terms to better understand their combined impact on success.

The regression models indicate that the range of the proposed metrics collectively predicts 75.6\% and 90.7\% of the generalized success rate in the two experiments, respectively. Interactions among the pathlength and jerk metrics are the most significant factors for estimating the generalized success rate in both experiments, as observed in the task success rate models. Details of the regression models and the analysis are provided in Appendix~\ref{appendix:reg_general}.

%% file: Text/discussion.tex
In this paper, we proposed a set of metrics to evaluate the quality of demonstrations prior to learning from demonstration (LfD). Specifically, we focused on consistency as a key measure of data quality and explored various motion characteristics to identify which metrics best predict learning performance. Unlike previous studies that artificially inject noise into data~\cite{brown2020better}, we evaluated our approach using real demonstration data collected from users with varying levels of robotics expertise. The experimental results clearly demonstrate that consistency in demonstrations significantly impacts both the success rate of learning and its generalization in different robot tasks.

Our clustering analysis revealed that inconsistent demonstrations are characterized by larger ranges and higher means in path length, jerk, and effort. In contrast, consistent demonstrations exhibit smaller ranges in these metrics, indicating smoother, more conservative, and less energy-intensive movements. This finding supports the notion that smoother and more controlled demonstrations provide better input for learning algorithms, reducing the likelihood of overfitting to noisy or suboptimal data.

Statistical analyses further highlight the significant contribution of consistency to both task success rates and generalization performance. These findings align with Pomerleau's work~\cite{pomerleau1988alvinn}, where consistent data provided clearer, more repeatable patterns for learning algorithms, leading to improved performance. Importantly, we observed that practice leads to significant improvements in consistency. However, the rate of improvement varied across metrics. For example, path length (in both Cartesian and joint spaces), manipulability, and curvature improved significantly after just one trial, whereas metrics like jerk, legibility, and effort required more practice to show meaningful improvement. This suggests that while some aspects of demonstrations can be refined quickly, others take time and experience to reach optimal levels. This opens up an exciting direction for future work, such as developing personalized training frameworks that track and adjust for individual user progress across specific metrics. By monitoring these metrics, we could avoid overtraining, which might otherwise diminish user performance, and ensure the highest quality demonstrations from each participant. Researchers have attempted to implement training curricula for users~\cite{ajaykumar2023curricula, sakr2020training, cakmak2014teaching}, but they did not provide a way to evaluate performance over trials, which the proposed metrics would help in achieving.

All of the proposed metrics can be computed during data collection, prior to running the learning algorithm, with the exception of legibility, which requires prior knowledge of task goals. Legibility becomes especially important in human-robot interaction scenarios, where the robot's actions need to be easily understood by humans~\cite{dragan2013legibility}. Thus, while legibility may not be necessary in all learning tasks, it plays a critical role in ensuring intuitive and seamless interaction in shared environments.

The correlation and regression analyses provided further insights into the metrics that are most influential in achieving high task success and generalization rates. They revealed that all metrics significantly impact the task learning and generalization success rate, either directly or through interaction with each other. However, the nature of the task influences the relative importance of each metric. For example, in the first experiment (button-pressing task in a constrained space), minimizing pathlength in joint space was essential to avoid self-collision and collision with the experimental setup. Orientation also played a key role in the button-pressing task, as the angle of approach was critical to successful execution. Interestingly, effort and jerk were highly correlated with pathlength and orientation, as shown in Fig.~\ref{fig:corr_heatmap}. Thus, their interactions have a high impact on the success rate, as shown in Table~\ref{tab:exp1_regress_task}. Notably, even though these metrics had the highest impact, the influence of the other metrics cannot be ignored. When we reran the regression using only these four metrics and ignored the others, only 45\% of the variance in the task success rate was captured, highlighting the importance of a comprehensive evaluation. These findings suggest that researchers should prioritize the selection of the most consistent demonstrations across all metrics to achieve high performance in both learning and generalization. 

In the second experiment (pick-and-place task), pathlength and jerk again emerged as critical metrics for both task and generalization success. The consistency of these findings across experiments highlights the robustness of these metrics in predicting learning performance. However, we also observed differences in the role of other metrics. For instance, the range of effort was more highly negatively correlated with success in the first experiment than the second one. Conversely, the range of distance to joint limits had a stronger negative correlation with success in the second experiment. These findings suggest that the specific nature of the task and its constraints influence which metrics are most important for optimizing learning and generalization.

Our proposed approach enables the evaluation of demonstration quality and consistency without the need to train the learning model, which sets it apart from prior work~\cite{sakr2022quantifying, gandhi2023eliciting}. Moreover, the proposed metrics implicitly capture inconsistencies in the solution strategy. For example, in the first experiment, users initially pushed the robot's shoulder joint to its limit to avoid collisions with a box. With practice, users discovered smoother and more efficient ways of manoeuvring around the box, as seen in Fig.\ref{fig:exp_good}. This pattern places the first trial in the inconsistent group, while subsequent trials fall into the consistent group. This suggests a promising direction for future work in active learning, where users could be guided to provide more consistent and higher-quality demonstrations, as explored in~\cite{gandhi2023eliciting}. While their approach requires retraining the policy and evaluating performance to measure compatibility, our method reveals inconsistencies and offers insights into demonstration quality without the need for training, providing a more efficient and practical approach.

%% file: Text/conclusion.tex
This paper presents an extensive set of metrics for evaluating the quality of demonstrations in Learning from Demonstration (LfD) tasks, with a particular focus on consistency as a key determinant of success. By rigorously analyzing real demonstration data from users of varying expertise, we have demonstrated that consistency in demonstrations significantly influences both the task success rate and generalization performance in robot learning. Our findings show that smoother, more controlled demonstrations, as indicated by smaller ranges in pathlength and jerk in both Cartesian and joint spaces, lead to better learning outcomes and enhanced generalization. Furthermore, our approach allows for real-time evaluation of demonstration quality, making it possible to assess data consistency prior to the learning process, thus ensuring that only high-quality data is used for robot training. The results of two user studies, involving distinct tasks and robot platforms, underline the robustness of the proposed metrics. Key metrics like pathlength and jerk in both Cartesian and joint spaces emerged as critical predictors of learning success across both experiments, highlighting their importance for a wide range of LfD tasks. Additionally, our work emphasizes the role of task-specific factors in shaping the contribution of each metric, suggesting that personalized training protocols could further improve the quality of demonstrations. 
Overall, this work fills an important gap in LfD research by offering a practical, data-driven method to evaluate and ensure demonstration quality before learning. It opens new avenues for optimizing robot learning and generalization in real-world scenarios, empowering non-expert users to teach robots effectively while reducing the risk of poor performance due to suboptimal data. Future work will focus on integrating these metrics into adaptive training frameworks and exploring active learning strategies to guide users to provide more consistent and higher quality demonstrations.

%% file: main.bbl

%% file: Text/appendix.tex
\renewcommand{\thesection}{A.\arabic{section}} 
    \renewcommand{\thetable}{A\arabic{table}} 
    \setcounter{table}{0} 

\section{Legibility calculations}
\label{appendix: leg}

The formula for calculating the legibility score is shown in Table~\ref{tab:metrics}, where \( P(G_i | \xi_{1:t}) \) is the posterior probability of goal \( G_i \) given the trajectory segment \( \xi_{1:t} \). The probability is computed using the cost function for each potential goal, which calculates the likelihood of a trajectory for a given goal. It combines early differentiation and progress towards the goal as follows:

\begin{equation}
C(\xi, G) = w_1 \left( \frac{1}{D_{\text{early}}} \right) + w_2 \left( \frac{1}{P_{\text{goal}}} \right)
\end{equation}

where \( D_{\text{early}} \) is the early differentiation score, \( P_{\text{goal}} \) is the progress towards the goal, and \( w_1 \) and \( w_2 \) are the weights for each term. Early differentiation measures how distinguishable the trajectory is from other potential goals early in the motion as follows:

\begin{equation}
D_{\text{early}} = \frac{1}{n} \sum_{i=1}^{n} \| \xi_{1:t} - G \|
\end{equation}

where $n$ is the number of potential goals, \( \xi_{1:t} \) is the early trajectory segment, and  \( G \) is the goal. Progress towards the goal measures how directly the trajectory moves towards the goal as follows:

\begin{equation}
P_{\text{goal}} = \frac{\sum_{i=1}^{n} \max(\text{proj}(\vec{\xi_i}, \vec{G}), 0)}{\|G - \xi_0\|}
\end{equation}

where \( \vec{\xi_i} \) is the vector segment of the trajectory, \( \vec{G} \) is the vector towards the goal, and \( \xi_0 \) is the starting point of the trajectory.

\section{Regression analysis of task success rate}
\label{appendix:reg_task}

Initially, linear regression was employed, but it captured only a small variation of the success rate using linear relationships with the predictors. Consequently, stepwise regression was applied to consider interactions among the predictors and their quadratic terms. Additionally, we standardized the predictors and outcome values to facilitate the interpretation of the main effects and interaction terms.

\begin{table*}[t]
\centering
\caption{Regression Model detailing the relation between the range of the proposed metrics and the task success rate in the first experiment. $x_1$ represents the range of pathlength in Cartesian space, $x_2$ represents the range of pathlength in joint space, $x_3$ represents the range of path orientation length, $x_4$ represents the range of jerk in Cartesian space, $x_5$ represents the range of jerk in joint space, $x_6$ represents the range of manipulability, $x_7$ represents the range of the distance to joint limits, $x_8$ represents the range of the demonstration curvature, $x_9$ represents the range of joint effort, $x_{10}$ represents the range of the legibility and $x_{11}$ represents the consistency.}
\begin{tabularx}{\textwidth}{llllX}
\toprule
\textbf{Coefficient} & \textbf{Estimate} & \textbf{SE} & \textbf{t-Statistic} & \textbf{Visualization of coefficients and their standard errors } \\
\midrule
(Intercept) & -0.5499 & 0.1092 & -5.0347 & \includegraphics[width=\linewidth, height=0.3cm]{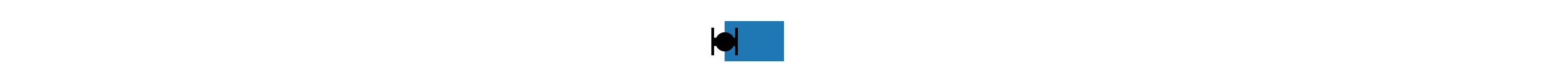} \\
Range of pathlength in joint space$(x_2)$ & \textbf{-3.8873} & 0.7807 & -4.9794 & \includegraphics[width=\linewidth, height=0.3cm]{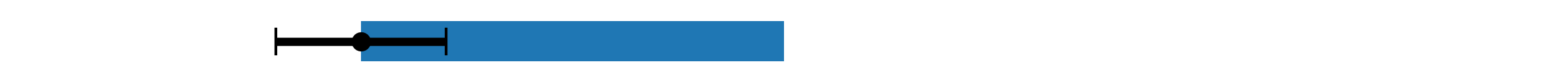} \\
Range of path orientation length$(x_3)$ & \textbf{2.9233} & 0.6930 & 4.2184 & \includegraphics[width=\linewidth, height=0.3cm]{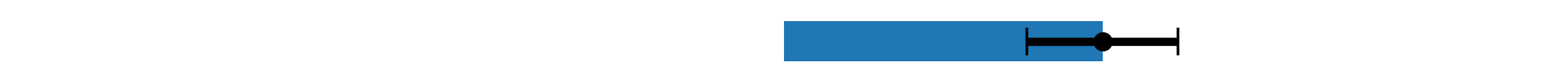} \\
Range of jerk in Cartesian space$(x_4)$ & -0.2674 & 0.0884 & -3.0254 & \includegraphics[width=\linewidth, height=0.3cm]{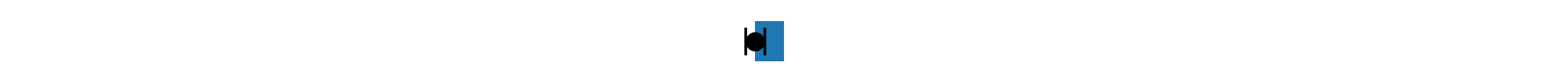} \\
Range of jerk in joint space$(x_5)$ & -0.3258 & 0.1448 & -2.2505 & \includegraphics[width=\linewidth, height=0.3cm]{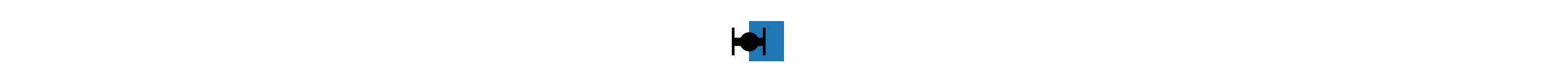} \\
Range of distance to joint limit$(x_7)$ & -0.1508 & 0.0612 & -2.4649 & \includegraphics[width=\linewidth, height=0.3cm]{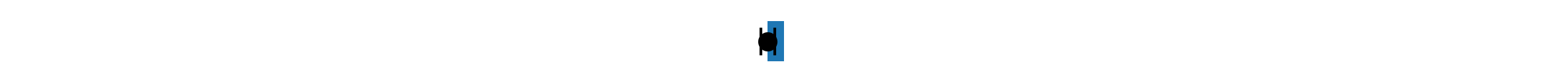} \\
Range of curvature$(x_8)$ & -0.5086 & 0.1198 & -4.2468 & \includegraphics[width=\linewidth, height=0.3cm]{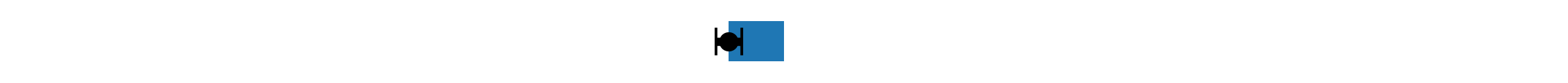} \\
Range of legibility$(x_{10})$ & 0.2706 & 0.0595 & 4.5450 & \includegraphics[width=\linewidth, height=0.3cm]{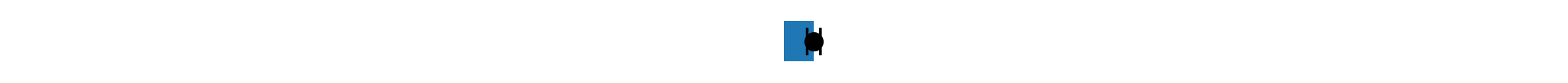} \\
Consistency$(x_{11})$ & 0.7208 & 0.1574 & 4.5795 & \includegraphics[width=\linewidth, height=0.3cm]{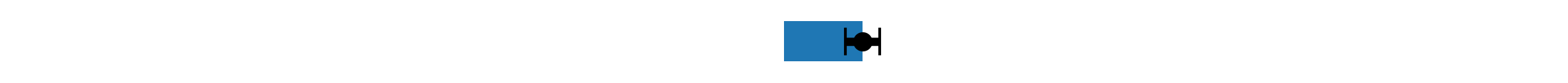} \\
$x_3:x_4$ & -0.4071 & 0.1274 & -3.1964 & \includegraphics[width=\linewidth, height=0.3cm]{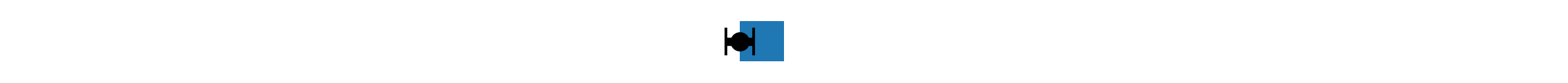} \\
$x_1:x_5$ & 0.5346 & 0.2232 & 2.3950 & \includegraphics[width=\linewidth, height=0.3cm]{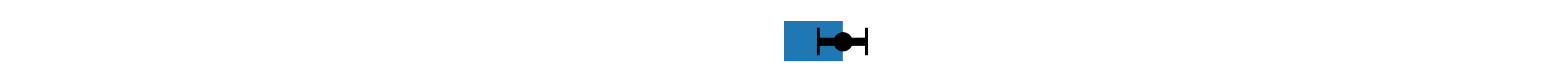} \\
$x_2:x_5$ & \textbf{-5.1903} & 1.1900 & -4.3617 & \includegraphics[width=\linewidth, height=0.3cm]{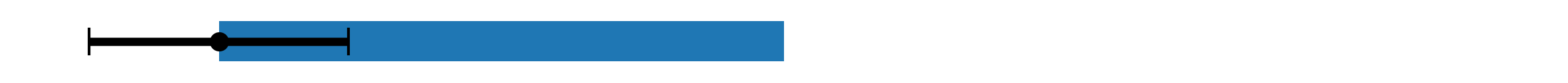} \\
$x_3:x_5$ & \textbf{4.6017} & 1.0848 & 4.2420 & \includegraphics[width=\linewidth, height=0.3cm]{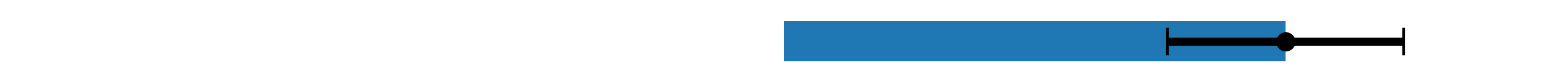} \\
$x_4:x_5$ & -0.4149 & 0.1721 & -2.4112 & \includegraphics[width=\linewidth, height=0.3cm]{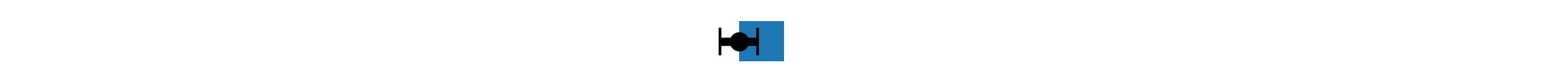} \\
$x_4:x_6$ & 0.3448 & 0.0848 & 4.0686 & \includegraphics[width=\linewidth, height=0.3cm]{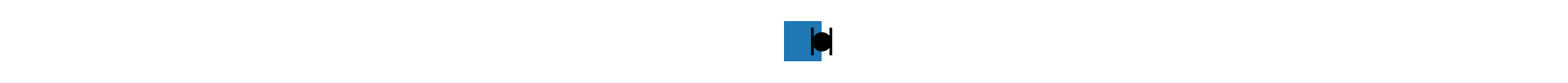} \\
$x_5:x_6$ & -0.6147 & 0.1307 & -4.7043 & \includegraphics[width=\linewidth, height=0.3cm]{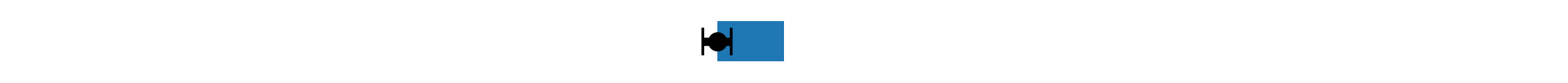} \\
$x_2:x_7$ & -0.2345 & 0.0641 & -3.6568 & \includegraphics[width=\linewidth, height=0.3cm]{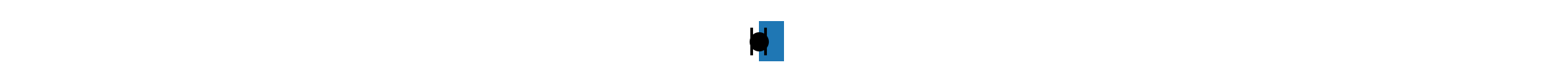} \\
$x_1:x_8$ & -0.9810 & 0.3257 & -3.0124 & \includegraphics[width=\linewidth, height=0.3cm]{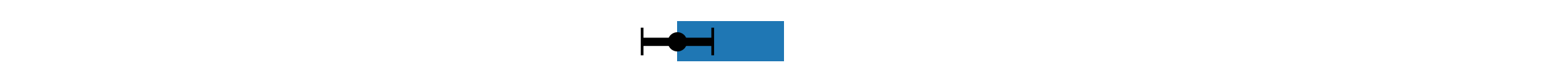} \\
$x_2:x_9$ & \textbf{3.6332} & 0.7405 & 4.9061 & \includegraphics[width=\linewidth, height=0.3cm]{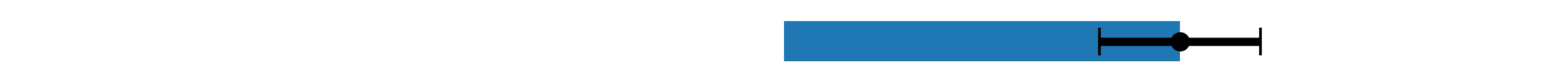} \\
$x_3:x_9$ & \textbf{-3.1695} & 0.7050 & -4.4955 & \includegraphics[width=\linewidth, height=0.3cm]{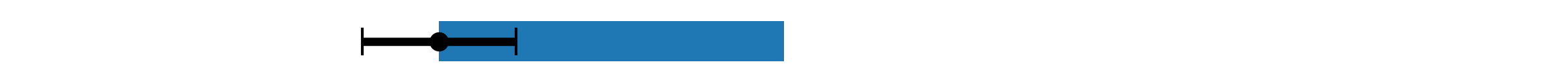} \\
$x_6:x_9$ & 0.2374 & 0.0628 & 3.7817 & \includegraphics[width=\linewidth, height=0.3cm]{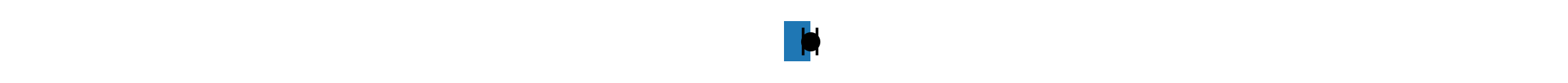} \\
$x_8:x_9$ & -0.3518 & 0.1390 & -2.5321 & \includegraphics[width=\linewidth, height=0.3cm]{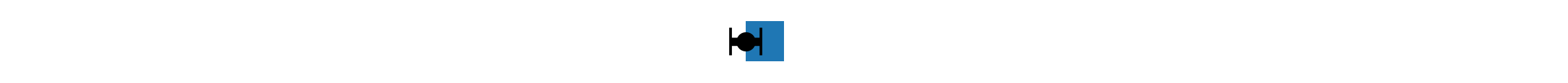} \\
$x_4:x_{10}$ & 0.2371 & 0.0509 & 4.6618 & \includegraphics[width=\linewidth, height=0.3cm]{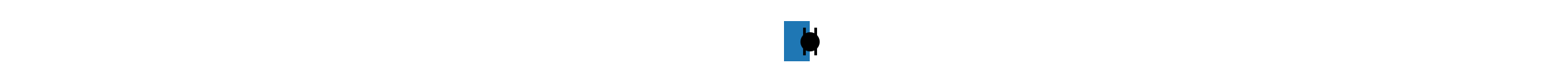} \\
$x_6:x_{10}$ & -0.2263 & 0.0754 & -3.0016 & \includegraphics[width=\linewidth, height=0.3cm]{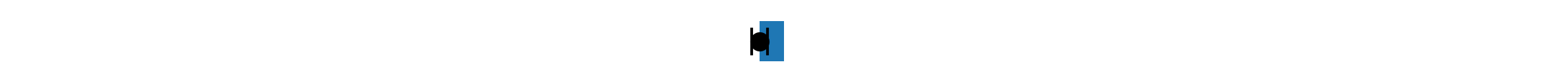} \\
$x_4^2$ & 0.2007 & 0.0502 & 3.9971 & \includegraphics[width=\linewidth, height=0.3cm]{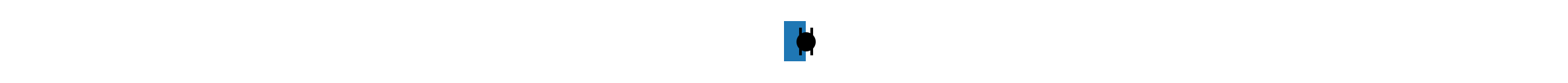} \\
$x_5^2$ & 0.5752 & 0.1976 & 2.9117 & \includegraphics[width=\linewidth, height=0.3cm]{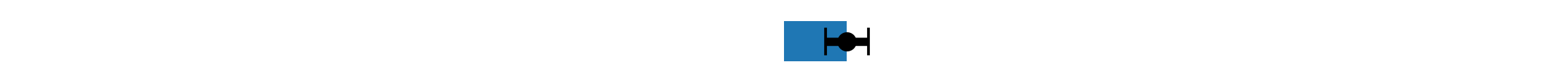} \\
\bottomrule
\end{tabularx}

\begin{tabularx}{\textwidth}{ll}
\textbf{Number of observations}: & 144 \\
\textbf{Error degrees of freedom}: & 118 \\
\textbf{Root Mean Squared Error}: & 0.601 \\
\textbf{R-squared}: & 0.702 \\
\textbf{Adjusted R-Squared}: & 0.639 \\
\textbf{F-statistic vs. constant model}: & 11.1 \\
\textbf{p-value}: & 9.14e-21 \\
\bottomrule
\end{tabularx}

\label{tab:exp1_regress_task}
\end{table*}

\begin{table*}[t]
\centering
\caption{Regression Model detailing the relation between the range of the proposed metrics and the task success rate in the second experiment. $x_1$ represents the range of pathlength in Cartesian space, $x_2$ represents the range of pathlength in joint space, $x_3$ represents the range of path orientation length, $x_4$ represents the range of jerk in Cartesian space, $x_5$ represents the range of jerk in joint space, $x_6$ represents the range of manipulability, $x_7$ represents the range of the distance to joint limits, $x_8$ represents the range of the demonstration curvature, $x_9$ represents the range of joint effort, and $x_{10}$ represents the consistency.}
\begin{tabularx}{\textwidth}{llllX}
\toprule
\textbf{Coefficient} & \textbf{Estimate} & \textbf{SE} & \textbf{t-Statistic} & \textbf{Visualization of coefficients and their standard errors} \\
\midrule
Intercept & -0.011078 & 0.090473 & -0.12244 & \includegraphics[width=\linewidth, height=0.3cm]{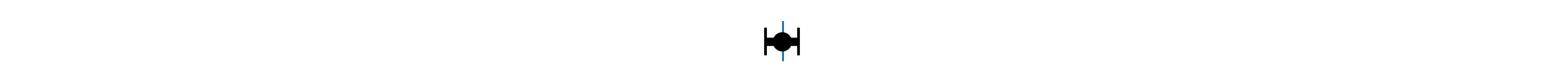} \\
Range of pathlength in Cartesian space ($x_1$) & -1.1226 & 0.30021 & -3.7392 & \includegraphics[width=\linewidth, height=0.3cm]{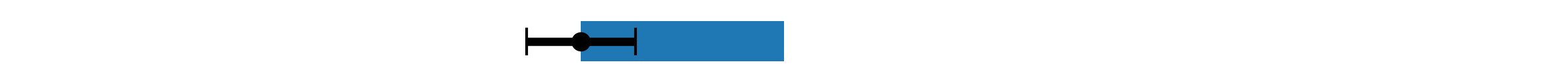} \\
Range of pathlength in joint space ($x_2$) & 0.85762 & 0.35571 & 2.411 & \includegraphics[width=\linewidth, height=0.3cm]{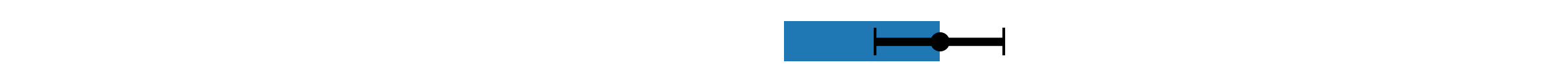} \\
Range of jerk in joint space ($x_5$) & -0.94164 & 0.2167 & -4.3454 & \includegraphics[width=\linewidth, height=0.3cm]{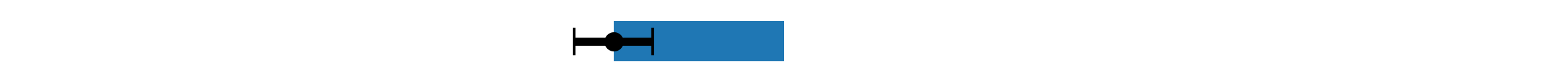} \\
$x_2:x_3$ & \textbf{3.135} & 0.57773 & 5.4264 & \includegraphics[width=\linewidth, height=0.3cm]{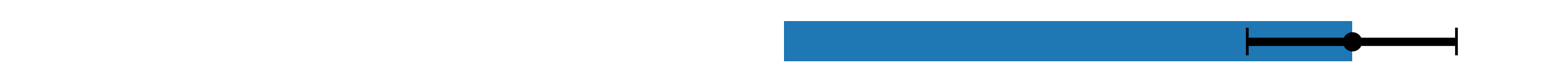} \\
$x_1:x_4$ & \textbf{-2.9919} & 0.52756 & -5.6711 & \includegraphics[width=\linewidth, height=0.3cm]{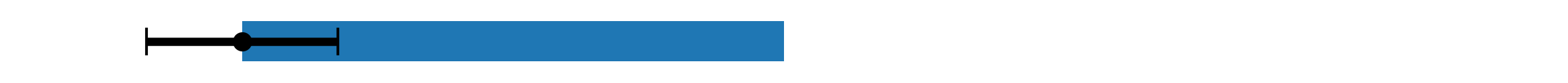} \\
$x_3:x_5$ & \textbf{-3.1553} & 0.68183 & -4.6277 & \includegraphics[width=\linewidth, height=0.3cm]{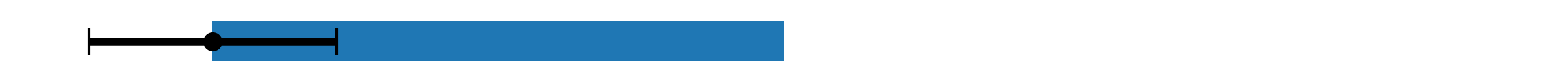} \\
$x_4:x_5$ & \textbf{2.4976} & 0.61275 & 4.0761 & \includegraphics[width=\linewidth, height=0.3cm]{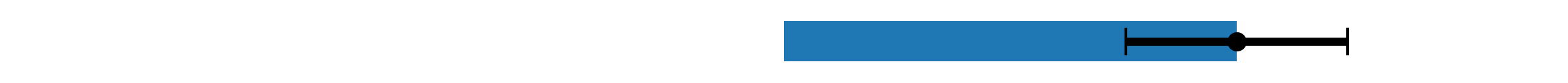} \\
$x_1:x_6$ & -1.3935 & 0.28438 & -4.9002 & \includegraphics[width=\linewidth, height=0.3cm]{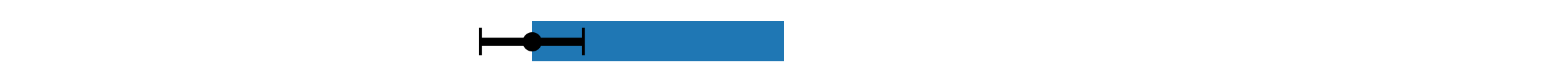} \\
$x_2:x_6$ & 1.6134 & 0.29548 & 5.4603 & \includegraphics[width=\linewidth, height=0.3cm]{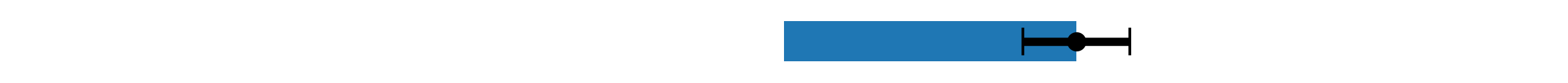} \\
$x_4:x_8$ & -1.2869 & 0.29361 & -4.3828 & \includegraphics[width=\linewidth, height=0.3cm]{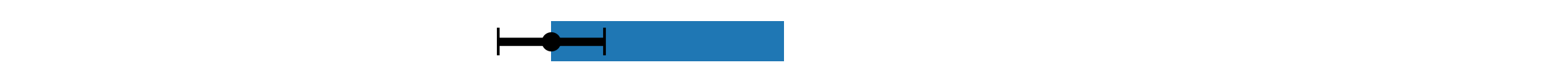} \\
$x_6:x_8$ & -1.0353 & 0.21373 & -4.8443 & \includegraphics[width=\linewidth, height=0.3cm]{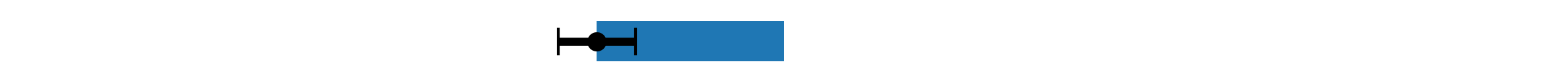} \\
$x_1:x_9$ & 1.0596 & 0.27262 & 3.8867 & \includegraphics[width=\linewidth, height=0.3cm]{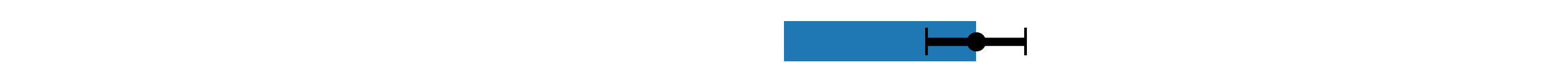} \\
$x_2:x_9$ & -0.99994 & 0.27296 & -3.6633 & \includegraphics[width=\linewidth, height=0.3cm]{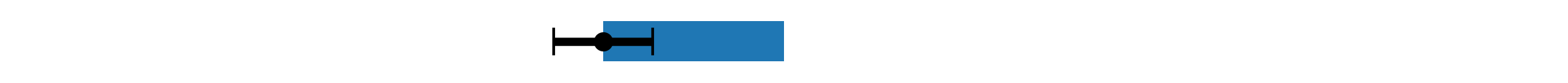} \\
$x_8:x_9$ & 0.92439 & 0.15412 & 5.998 & \includegraphics[width=\linewidth, height=0.3cm]{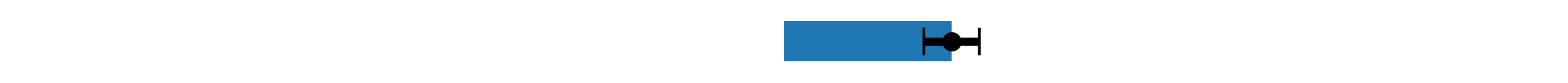} \\
$x_4:x_{10}$ & -0.48341 & 0.18134 & -2.6657 & \includegraphics[width=\linewidth, height=0.3cm]{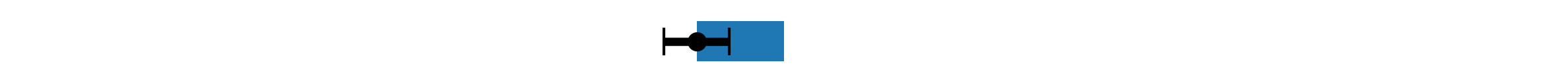} \\
\bottomrule
\end{tabularx}

\begin{tabularx}{\textwidth}{ll}
\textbf{Number of observations}: & 60 \\
\textbf{Error degrees of freedom}: & 44 \\
\textbf{Root Mean Squared Error}: & 0.392 \\
\textbf{R-squared}: & 0.886 \\
\textbf{Adjusted R-Squared}: & 0.847 \\
\textbf{F-statistic vs. constant model}: & 22.7 \\
\textbf{p-value}: & 6.95e-16 \\
\bottomrule
\end{tabularx}

\label{tab:exp2_task_regress}
\end{table*}

Table~\ref{tab:exp1_regress_task} shows the resulting regression model that represents the relationship between the range of the metrics and the task success rate. The model is significant and captures 70.2\% of the variance in the success rate. The model shows a significant linear relationship with eight predictors that are the range of: pathlength in joint space$(x_2)$, path orientation length $(x_3)$, jerk in Cartesian space $(x_4)$, jerk in joint space $(x_5)$, distance to joint limit $(x_7)$, curvature$(x_8)$, legibility$(x_{10})$, and the consistency $(x_{11})$. While the other metrics exhibit a more complex relationship with the success rate through interaction and quadratic terms, indicating the importance of evaluating the metrics together, rather than focusing on any single metric. 

The visualization in Table~\ref{tab:exp1_regress_task} highlights the coefficients and standard errors of the significant terms. Notably, range of the pathlength in joint space ($x_2$), orientation pathlength ($x_3$), and their interaction with jerk in joint space ($x_5$) and effort ($x_9$) are crucial for determining success, given their higher coefficient values. The negative coefficient of ($x_2$) suggests that shorter, more consistent paths reduce unnecessary movements, thus lowering the likelihood of errors~\cite{osa2018algorithmic}. Variability in both pathlength and jerk in joint space reflects inefficient, erratic movements, detrimental to precision-based tasks, which is why their interaction term negatively impacts success rate. Conversely, the positive interaction between pathlength in joint space and effort suggests that aligning effort with path variability improves control and adaptation, enhancing performance.

The range of orientation pathlength ($x_3$) has a positive effect on the success rate, indicating the importance of flexible orientation adjustments to complete the task, especially in high-constraint environments. Additionally, its interaction with jerk in joint space positively influences success, suggesting that quick, even jerky, orientation adjustments are beneficial in such contexts. However, its interaction with effort shows a negative coefficient, indicating that excessive variability in both orientation pathlength and effort reduces success. This suggests that high, consistent effort associated with orientation adjustments may impede performance.

Similarly, a regression model was developed for the second experiment, examining the relationship between task performance success rate and the range of metrics. Table~\ref{tab:exp2_task_regress} illustrates the model and provides a visualization of the coefficients and standard errors of significant terms. The model is significant, capturing 88.6\% of the variance in success rate. Key interactions involve pathlength in Cartesian space ($x_1$), pathlength in joint space ($x_2$), orientation pathlength ($x_3$), jerk in Cartesian space ($x_4$), and jerk in joint space ($x_5$), with higher coefficient values highlighting their importance in estimating success rate. The positive interaction between pathlength in joint space and orientation pathlength suggests that flexible orientation adjustments, alongside variability in joint movements, enhance performance in dynamic tasks. However, the negative interaction between pathlength and jerk in Cartesian space indicates that excessive variability in both metrics can hinder task performance. Moreover, interactions involving jerk in joint space with both orientation pathlength and jerk in Cartesian space exhibit contrasting effects: the former negatively impacts success, reflecting that erratic joint movements combined with orientation changes reduce performance, while the latter positively impacts success, suggesting that coordinated variability in jerk across joint and Cartesian spaces improves adaptability, leading to better task outcomes.

\section{Regression analysis of generalized success rate}
\label{appendix:reg_general}

Similar regression analysis was conducted to explore the relationship between the range of the metrics and the consistency label with the generalized success rate for the two experiments data. Table~\ref{tab:gen_regress} presents the regression model detailing the relationship between the range of the proposed metrics and the generalized success rate of the button pressing experiment. The model is significant, explaining 75.6\% of the variance in the generalized success rate. Notably, the same metrics that were important in predicting the task success rate also contribute the most to the generalized success rate. These key metrics include the range of pathlength in joint space $(x_2)$, the range of path orientation length $(x_3)$, and their interaction with jerk in joint space $(x_5)$. This underscores the critical role of these metrics in predicting both task and generalized success.

\begin{table*}[t]
\centering
\caption{Regression Model detailing the relationship between the range of the proposed metrics and the generalized success rate in the first experiment. $x_1$ represents the range of pathlength in Cartesian space, $x_2$ represents the range of pathlength in joint space, $x_3$ represents the range of path orientation length, $x_4$ represents the range of jerk in Cartesian space, $x_5$ represents the range of jerk in joint space, $x_6$ represents the range of manipulability, $x_7$ represents the range of the distance to joint limits, $x_8$ represents the range of the demonstration curvature, $x_9$ represents the range of joint effort, $x_{10}$ represents the range of legibility and $x_{11}$ represents the consistency.}
\begin{tabularx}{\textwidth}{llllX}
\toprule
\textbf{Coefficient} & \textbf{Estimate} & \textbf{SE} & \textbf{t-Statistic} & \textbf{Visualization of coefficients and their standard errors} \\
\midrule
(Intercept)     &  0.27975   & 0.12896   &  2.1692   & \includegraphics[width=\linewidth, height=0.3cm]{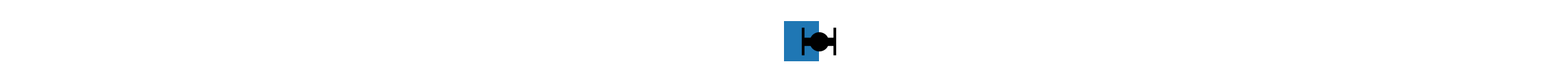} \\
Range of pathlength in joint space$(x_2)$           & \textbf{-3.5869}    & 0.7299    & -4.9143   & \includegraphics[width=\linewidth, height=0.3cm]{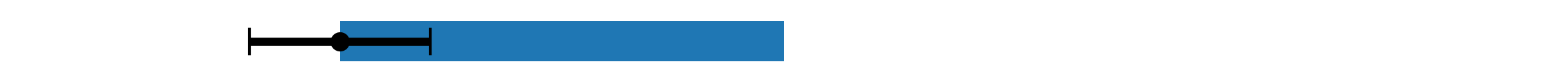} \\
Range of path orientation length$(x_3)$           &  \textbf{2.8648}    & 0.68134   &  4.2047   & \includegraphics[width=\linewidth, height=0.3cm]{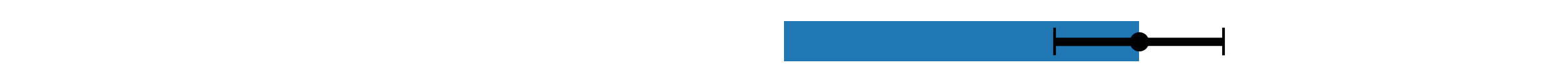} \\
Range of jerk in joint space$(x_5)$           & -0.49044   & 0.12662   & -3.8734   & \includegraphics[width=\linewidth, height=0.3cm]{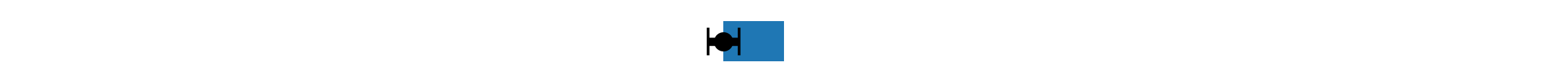} \\
Range of joint effort$(x_9)$           & -0.33265   & 0.11054   & -3.0092   & \includegraphics[width=\linewidth, height=0.3cm]{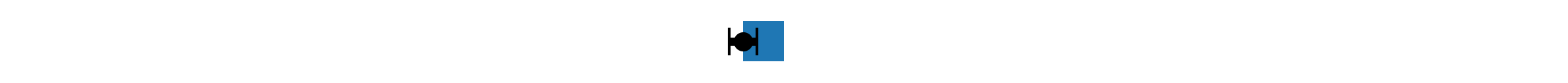} \\
Consistency$(x_{11})$        & -1.2349    & 0.32368   & -3.8152   & \includegraphics[width=\linewidth, height=0.3cm]{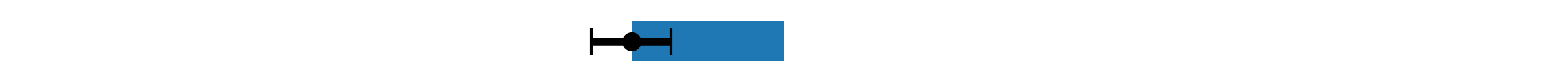} \\
$x_1:x_4$       &  0.82844   & 0.19886   &  4.1659   & \includegraphics[width=\linewidth, height=0.3cm]{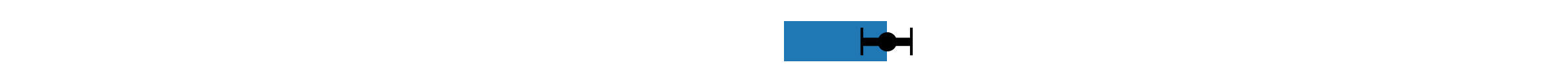} \\
$x_3:x_4$       & -0.90945   & 0.25697   & -3.5391   & \includegraphics[width=\linewidth, height=0.3cm]{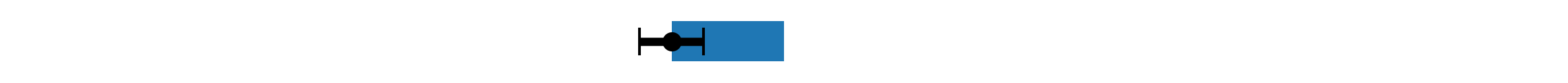} \\
$x_2:x_5$       & \textbf{-3.7961}    & 0.73839   & -5.141    & \includegraphics[width=\linewidth, height=0.3cm]{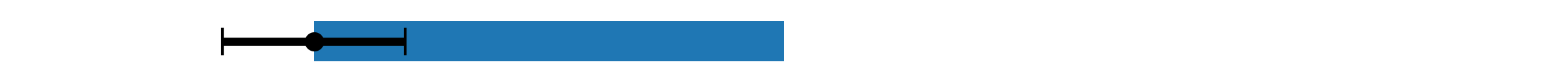} \\
$x_3:x_5$       &  \textbf{4.7715}    & 0.83905   &  5.6868   & \includegraphics[width=\linewidth, height=0.3cm]{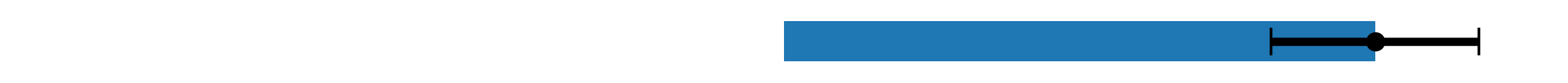} \\
$x_4:x_5$       & -0.35898   & 0.13353   & -2.6883   & \includegraphics[width=\linewidth, height=0.3cm]{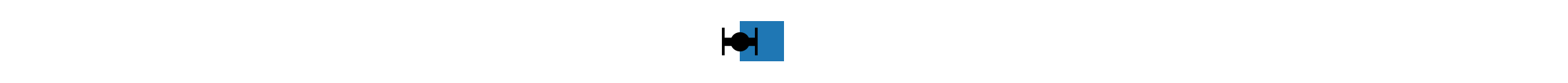} \\
$x_4:x_6$       &  0.1458    & 0.073164  &  1.9927   & \includegraphics[width=\linewidth, height=0.3cm]{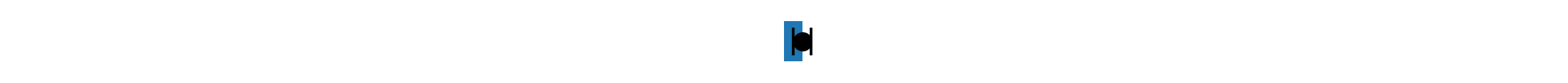} \\
$x_5:x_6$       & -0.86245   & 0.12318   & -7.0017   & \includegraphics[width=\linewidth, height=0.3cm]{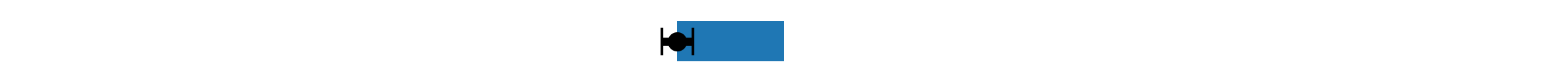} \\
$x_1:x_7$       &  1.3359    & 0.28081   &  4.7574   & \includegraphics[width=\linewidth, height=0.3cm]{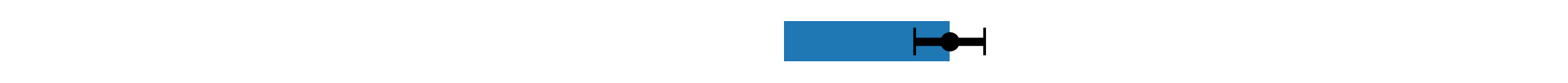} \\
$x_2:x_7$       & -0.63637   & 0.27687   & -2.2984   & \includegraphics[width=\linewidth, height=0.3cm]{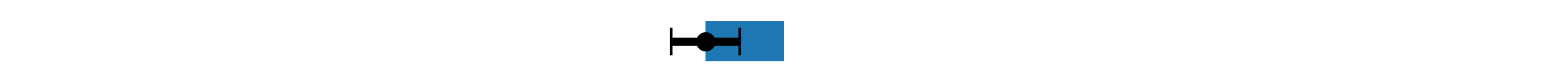} \\
$x_5:x_7$       &  0.27406   & 0.12249   &  2.2373   & \includegraphics[width=\linewidth, height=0.3cm]{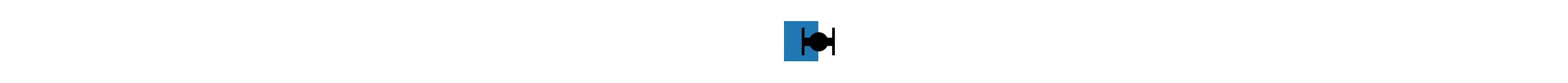} \\
$x_6:x_7$       & -0.29517   & 0.061987  & -4.7617   & \includegraphics[width=\linewidth, height=0.3cm]{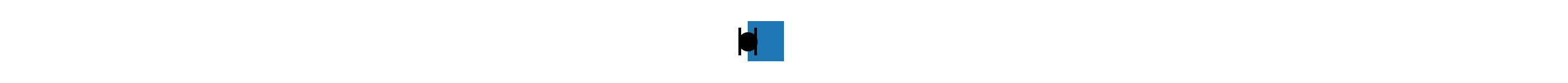} \\
$x_2:x_8$       &  1.6195    & 0.34553   &  4.687    & \includegraphics[width=\linewidth, height=0.3cm]{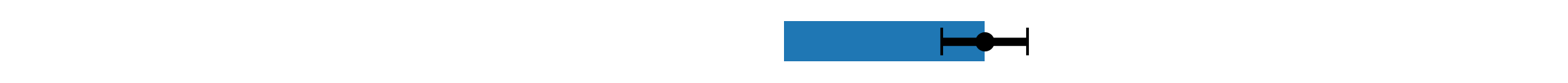} \\
$x_2:x_9$       &  1.4209    & 0.27381   &  5.1892   & \includegraphics[width=\linewidth, height=0.3cm]{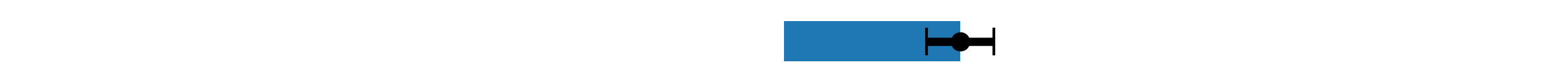} \\
$x_5:x_9$       & -0.66221   & 0.22121   & -2.9936   & \includegraphics[width=\linewidth, height=0.3cm]{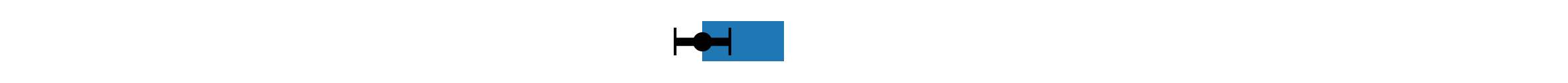} \\
$x_6:x_9$       &  0.44656   & 0.08497   &  5.2555   & \includegraphics[width=\linewidth, height=0.3cm]{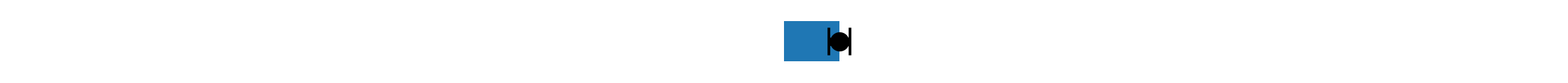} \\
$x_8:x_9$       & -0.74627   & 0.18647   & -4.0022   & \includegraphics[width=\linewidth, height=0.3cm]{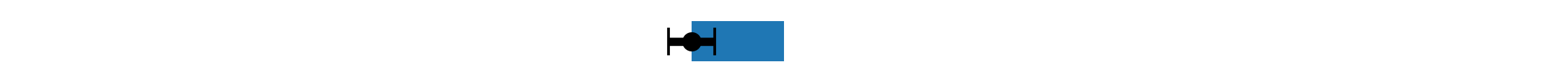} \\
$x_2:x_{10}$    & -0.40238   & 0.20063   & -2.0056   & \includegraphics[width=\linewidth, height=0.3cm]{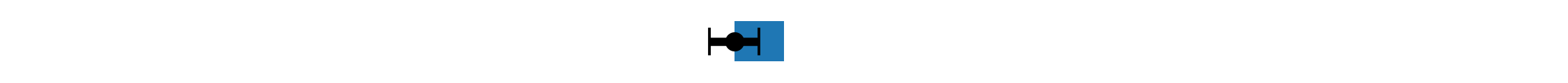} \\
$x_7:x_{10}$    & -0.36737   & 0.059606  & -6.1633   & \includegraphics[width=\linewidth, height=0.3cm]{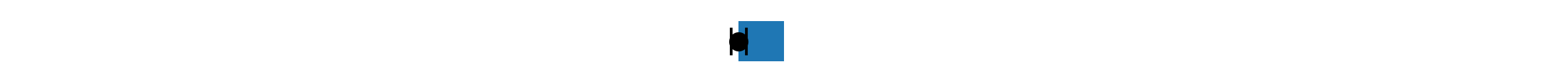} \\
$x_2:x_{11}$    &  1.1933    & 0.25787   &  4.6274   & \includegraphics[width=\linewidth, height=0.3cm]{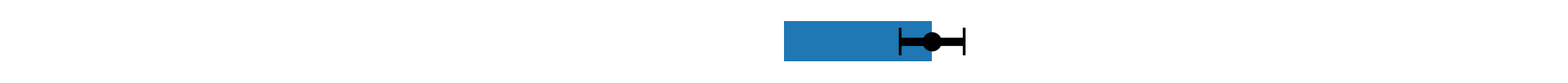} \\
$x_7:x_{11}$    &  0.93652   & 0.16325   &  5.7367   & \includegraphics[width=\linewidth, height=0.3cm]{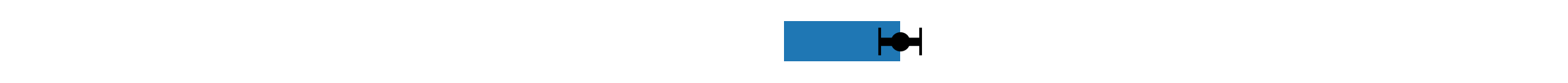} \\
$x_9:x_{11}$    &  0.46362   & 0.10807   &  4.2899   & \includegraphics[width=\linewidth, height=0.3cm]{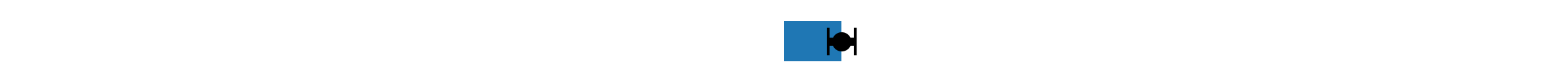} \\
$x_{10}:x_{11}$ & -0.44841   & 0.17632   & -2.5432   & \includegraphics[width=\linewidth, height=0.3cm]{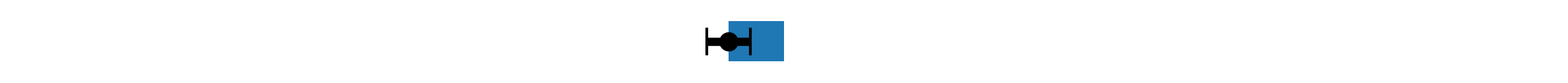} \\
\bottomrule
\end{tabularx}

\begin{tabularx}{\textwidth}{ll}
\textbf{Number of observations}: & 144 \\
\textbf{Error degrees of freedom}: & 116 \\
\textbf{Root Mean Squared Error}: & 0.549 \\
\textbf{R-squared}: & 0.756 \\
\textbf{Adjusted R-Squared}: & 0.699 \\
\textbf{F-statistic vs. constant model}: & 13.3 \\
\textbf{p-value}: & 2.43e-24 \\
\bottomrule
\end{tabularx}

\label{tab:gen_regress}
\end{table*}

In the pick and place experiment, the regression model (Table~\ref{tab:exp2_gen_regress}) explains 90.7\% of the variance in the generalized success rate. Key interactions involving pathlength in Cartesian space ($x_1$), pathlength in joint space ($x_2$), jerk in Cartesian space ($x_4$), and jerk in joint space ($x_5$) are among the most important terms, similar to the task performance model. This underscores the critical role of these metrics in predicting both task and generalized success. Consistency ($x_{10}$) also plays a critical role, positively affecting the generalized success rate, indicating that consistent demonstrations significantly improve the robot's ability to generalize.


\begin{table*}[t]
\centering
\caption{Regression Model detailing the relation between the range of the proposed metrics and the generalized success rate in the second experiment. $x_1$ represents the range of pathlength in Cartesian space, $x_2$ represents the range of pathlength in joint space, $x_3$ represents the range of path orientation length, $x_4$ represents the range of jerk in Cartesian space, $x_5$ represents the range of jerk in joint space, $x_6$ represents the range of manipulability, $x_7$ represents the range of the distance to joint limits, $x_8$ represents the range of the demonstration curvature, $x_9$ represents the range of joint effort, and $x_{10}$ represents the consistency.}
\begin{tabularx}{\textwidth}{llllX}
\toprule
\textbf{Coefficient} & \textbf{Estimate} & \textbf{SE} & \textbf{t-Statistic} & \textbf{Visualization of coefficients and their standard errors} \\
\midrule
Intercept & -2.1426 & 0.25463 & -8.4146 & \includegraphics[width=\linewidth, height=0.3cm]{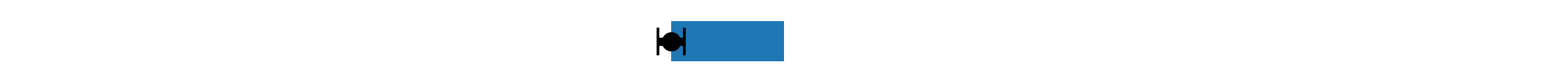} \\
Range of pathlength in joint space ($x_2$) & 1.5602 & 0.2692 & 5.7956 & \includegraphics[width=\linewidth, height=0.3cm]{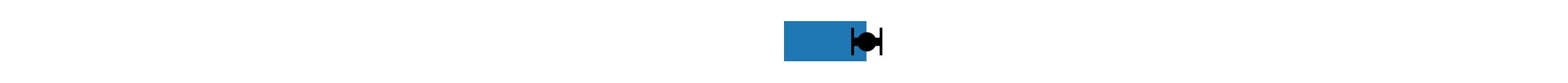} \\
Range of jerk in joint space ($x_5$) & -1.1614 & 0.2206 & -5.2647 & \includegraphics[width=\linewidth, height=0.3cm]{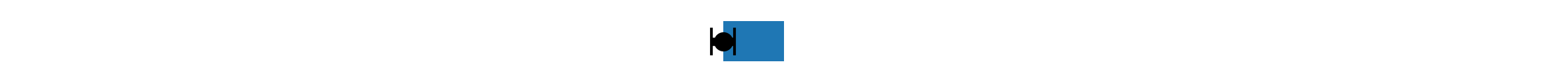} \\
Range of manipulability ($x_6$) & 0.46128 & 0.090664 & 5.0878 & \includegraphics[width=\linewidth, height=0.3cm]{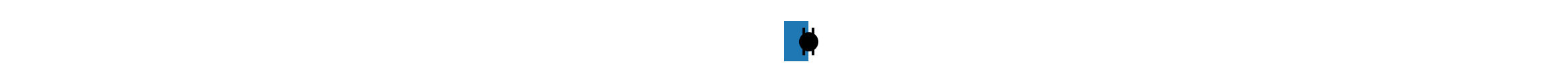} \\
Range of effort ($x_9$) & 0.45899 & 0.085092 & 5.394 & \includegraphics[width=\linewidth, height=0.3cm]{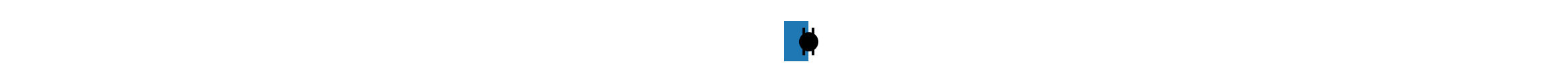} \\
Consistency ($x_{10}$) & \textbf{6.933} & 0.83754 & 8.2778 & \includegraphics[width=\linewidth, height=0.3cm]{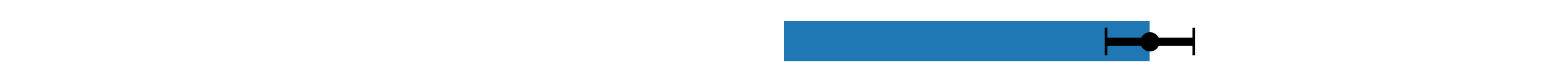} \\
$x_1:x_2$ & \textbf{11.712} & 1.1528 & 10.159 & \includegraphics[width=\linewidth, height=0.3cm]{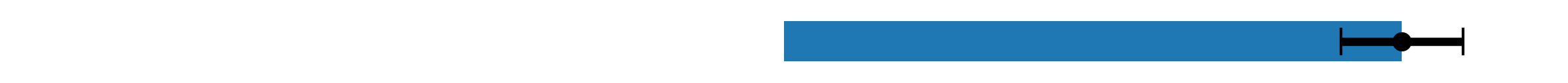} \\
$x_1:x_4$ & \textbf{-12.053} & 1.1215 & -10.748 & \includegraphics[width=\linewidth, height=0.3cm]{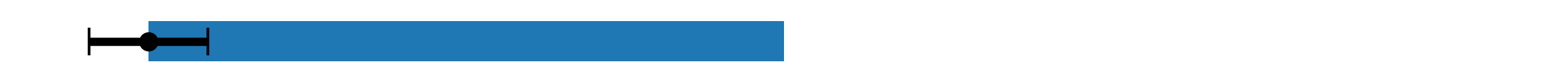} \\
$x_2:x_5$ & \textbf{-8.8461} & 0.89576 & -9.8755 & \includegraphics[width=\linewidth, height=0.3cm]{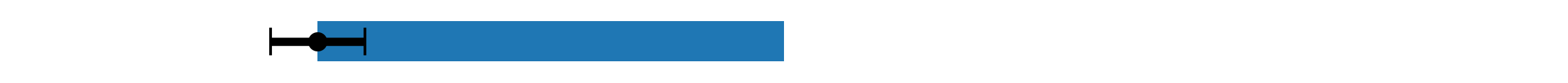} \\
$x_4:x_5$ & \textbf{8.4328} & 0.85875 & 9.8198 & \includegraphics[width=\linewidth, height=0.3cm]{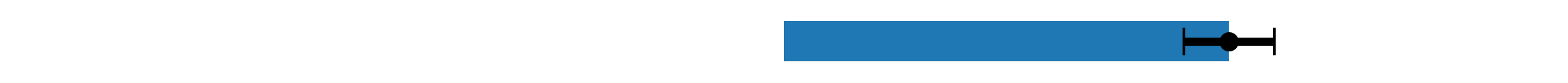} \\
$x_1:x_6$ & -1.649 & 0.27123 & -6.0799 & \includegraphics[width=\linewidth, height=0.3cm]{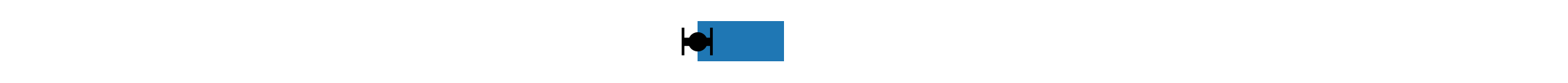} \\
$x_2:x_6$ & 2.1474 & 0.27391 & 7.8396 & \includegraphics[width=\linewidth, height=0.3cm]{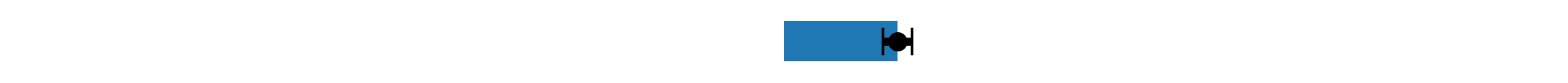} \\
$x_7:x_8$ & -0.22681 & 0.051082 & -4.4401 & \includegraphics[width=\linewidth, height=0.3cm]{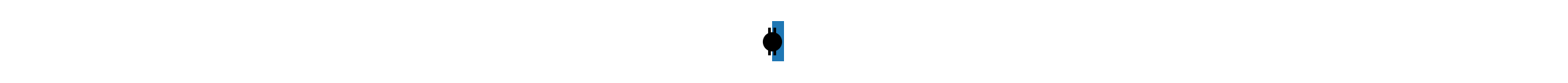} \\
$x_2:x_9$ & -4.2964 & 0.44981 & -9.5516 & \includegraphics[width=\linewidth, height=0.3cm]{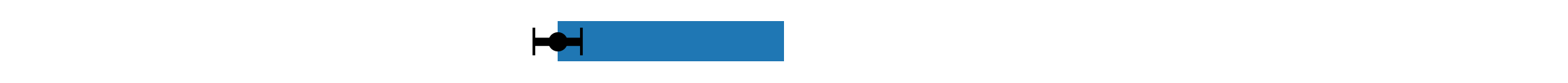} \\
$x_4:x_9$ & 4.946 & 0.49788 & 9.9341 & \includegraphics[width=\linewidth, height=0.3cm]{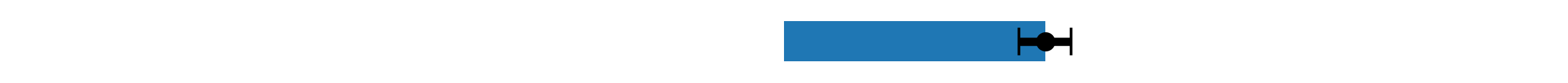} \\
$x_7:x_9$ & -0.28161 & 0.10056 & -2.8003 & \includegraphics[width=\linewidth, height=0.3cm]{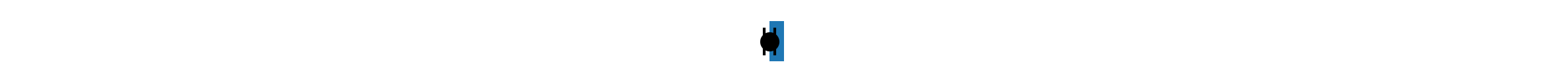} \\
$x_4:x_{10}$ & -4.5763 & 0.56098 & -8.1576 & \includegraphics[width=\linewidth, height=0.3cm]{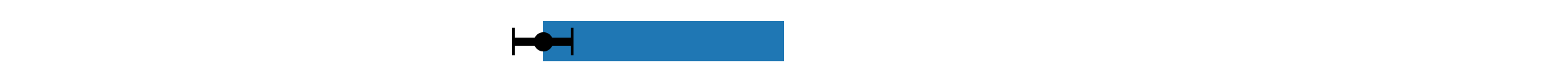} \\
$x_5:x_{10}$ & 1.6174 & 0.3304 & 4.8954 & \includegraphics[width=\linewidth, height=0.3cm]{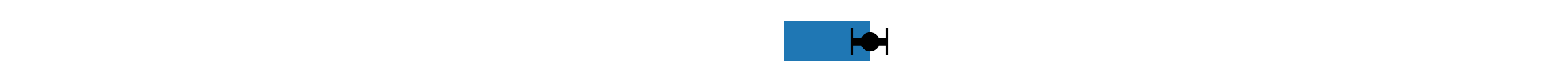} \\
$x_6:x_{10}$ & -0.69759 & 0.2299 & -3.0343 & \includegraphics[width=\linewidth, height=0.3cm]{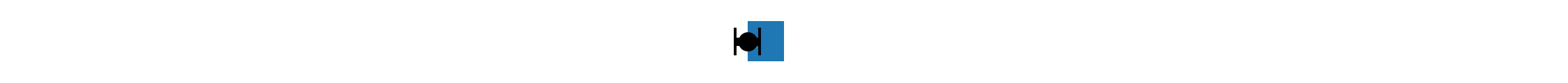} \\
$x_9:x_{10}$ & -0.74814 & 0.13176 & -5.6778 & \includegraphics[width=\linewidth, height=0.3cm]{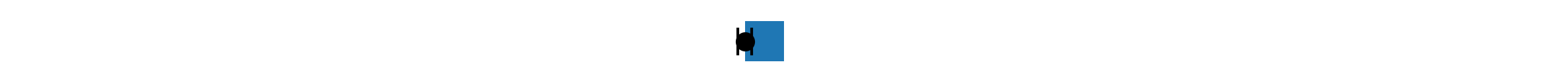} \\
\bottomrule
\end{tabularx}

\begin{tabularx}{\textwidth}{ll}
\textbf{Number of observations}: & 60 \\
\textbf{Error degrees of freedom}: & 40 \\
\textbf{Root Mean Squared Error}: & 0.371 \\
\textbf{R-squared}: & 0.907 \\
\textbf{Adjusted R-Squared}: & 0.862 \\
\textbf{F-statistic vs. constant model}: & 20.5 \\
\textbf{p-value}: & 6.37e-15 \\
\bottomrule
\end{tabularx}

\label{tab:exp2_gen_regress}
\end{table*}